\documentclass[letterpaper]{article} 
\usepackage{aaai2027}  
\usepackage[hyphens]{url}  
\usepackage{graphicx} 
\usepackage{natbib}  
\usepackage{caption} 
\usepackage[hyphens]{url}  
\usepackage{graphicx} 
\usepackage{natbib}  
\usepackage{caption} 
\usepackage{amssymb} 
\usepackage{amsmath} 
\usepackage{multirow} 
\usepackage{algorithm}
\usepackage{algorithmic}
\usepackage{booktabs}
\usepackage{multirow}
\usepackage[table]{xcolor}
\usepackage{array}

\definecolor{CJBlue}{RGB}{218,235,255}
\definecolor{CJBlueStrong}{RGB}{190,220,255}

\newcommand{\best}[1]{\textbf{#1}}
\newcommand{\second}[1]{\underline{#1}}
\newcommand{\cj}[1]{\cellcolor{CJBlueStrong}#1}
\newcommand{\cjb}[1]{\cellcolor{CJBlue}#1}
\usepackage{newfloat}
\usepackage{listings}
\DeclareCaptionStyle{ruled}{labelfont=normalfont,labelsep=colon,strut=off} 
\floatstyle{ruled}
\newfloat{listing}{tb}{lst}{}
\floatname{listing}{Listing}

\usepackage{booktabs}

\usepackage{algorithm}
\usepackage{algorithmic}

\usepackage{newfloat}
\usepackage{listings}
\DeclareCaptionStyle{ruled}{labelfont=normalfont,labelsep=colon,strut=off} 
\floatstyle{ruled}
\newfloat{listing}{tb}{lst}{}
\floatname{listing}{Listing}

\usepackage{booktabs}

\title{ChangeJEPA: Predict Before You Compare for Remote Sensing Change Detection}

\author {
    Buddhi Wijenayake,
    Dhananjaya Jayasundara,
    Vishal M. Patel
}
\affiliations {
    Johns Hopkins University\\
    \{bwarnas1,vjayasu1,vpatel36\}@jhu.edu
}

\begin{document}

\maketitle

\begin{abstract}
Remote sensing change detection is typically formulated as Siamese feature comparison, where two images are encoded independently and a decoder determines which discrepancies correspond to meaningful change. This is challenging because the encoders preserve both true scene changes and nuisance variations such as illumination, seasonal, radiometric, and geometric differences. We introduce ChangeJEPA, a bitemporal joint-embedding predictive architecture that reformulates change detection as latent post-state prediction. Given a pre-change image and partial post-change evidence, ChangeJEPA predicts the full post-change representation under two complementary conditions: a prior hypothesis derived from the pre-change scene and an updated hypothesis conditioned on observed post-change context. A joint-embedding predictive loss aligns both hypotheses with the target latent representation, encouraging semantic consistency while suppressing transient appearance variations. Change is then inferred from the disagreement among the pre-change representation, the observed post-change target, and the predicted latent states, capturing complementary evidence of change, prior surprise, contextual update, and prediction reliability. Across SYSU-CD, LEVIR-CD+, and CDD, ChangeJEPA consistently outperforms its decoder matched Siamese baseline by up to 1.51 IoU points and achieves the best IoU and F1-score on all three benchmarks, demonstrating the advantage of latent post-state prediction over direct feature comparison.
\begin{links}
    \link{Code}{https://github.com/Buddhi19/ChangeJEPA.git}
    \link{Project Page}{https://buddhi19.github.io/ChangeJEPA}
\end{links}
\end{abstract}

\section{Introduction}
\label{sec:intro}
\begin{figure}[ht]
    \centering
    \includegraphics[width=1\linewidth]{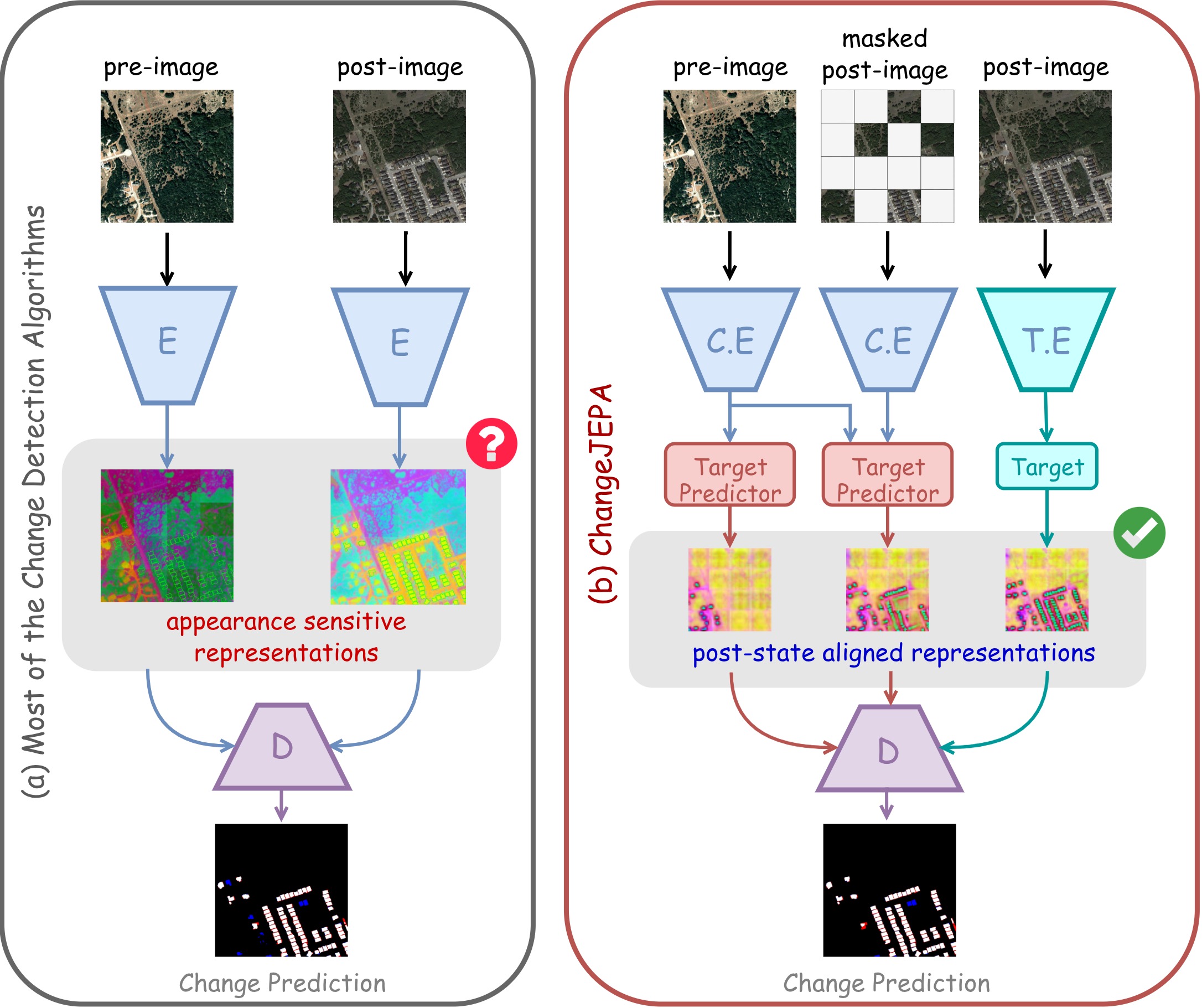}
\caption{\textbf{From appearance-sensitive to post-state-aligned latent representations.}
\textbf{Left:} Features from the 23rd block of a DINO-based Siamese encoder remain sensitive to bitemporal appearance variation, entangling genuine change with nuisance differences.
\textbf{Right:} ChangeJEPA aligns prior and updated predictions with the complete post-change target, yielding structured post-state evidence for change detection.
Here, E denotes the Siamese encoder, C.E.\ the context encoder, T.E.\ the target encoder, and D the decoder.}
\label{fig:sell}
\end{figure}
Remote sensing change detection (RSCD) aims to identify surface changes within the same geographical region from images acquired at different times \cite{chengChangeDetectionMethods2024a, maChangeDetectionRemote2019}. It is a central problem in Earth observation, supporting urban expansion analysis, land-cover monitoring, ecological assessment, resource management, disaster response , and other applications \cite{willisRemoteSensingChange2015, washayaCoherenceChangeDetectionSentinel12018}.

With the advancement of deep learning, RSCD has moved from hand-crafted comparison operators towards data-driven representation learning \cite{9136674}. Convolutional Siamese networks first provided an effective formulation by extracting bitemporal features and comparing them for dense prediction \cite{8022932,8451652}. More recently, Vision Transformer and visual state-space architectures have
strengthened long-range dependency modelling, global context aggregation, and
multi-scale bitemporal interaction, producing increasingly accurate and
spatially coherent change maps
\cite{changeformer,changemamba,BIT,yangQuadraticLinearTimeChange2026}.

Despite this architectural progress, most RSCD models remain task-specific
systems whose representations are learned primarily from annotated bitemporal
image pairs. This dependence has prompted growing interest in transferring
the broader visual knowledge of foundation models to change detection \cite{10438490}. CLIP-based methods such as ChangeCLIP adapt vision-language representations for change detection \cite{changeclip}, while SAM-based approaches such as AnyChange exploit segmentation foundation models for zero-shot or weakly supervised change localisation \cite{zhengSegmentAnyChange2024}. Self-supervised and generative foundation models have also begun to influence the field \cite{9844015,11730595}. For instance, ChangeDINO transfers DINO-derived semantic representations into a multiscale Siamese detector \cite{changeDINO}, whereas Changen2 synthesises multi-temporal change data to improve generalisation and reduce annotation dependence \cite{10713915}. These studies demonstrate the value of foundation-model knowledge for RSCD.

However, the dominant RSCD paradigm still relies on the same underlying principle: two observations are encoded independently, their representations are compared, and a decoder infers change from the resulting bitemporal discrepancy \cite{8022932,8451652,BIT,changemamba,ahnGeneralizableDisasterDamage2025,zhangUniChangeUnifyingChange2026}. As illustrated in Fig.~\ref{fig:sell}, these discrepancies are inherently ambiguous because nuisance factors such as illumination, shadows, seasonal variation, viewing geometry, radiometric differences, and registration errors can produce responses similar to genuine semantic change \cite{chengChangeDetectionMethods2024a,zangFeatureSpectrumLearning2025,changeDINO}. Consequently, the decoder must learn to separate true changes from pseudo-changes after they have already become entangled in the feature representations \cite{BIT,changemamba,changeDINO,11450773}. Thus, the fundamental limitation is not simply encoder or decoder capacity, but the ambiguity of direct bitemporal feature comparison, which identifies where images differ without indicating whether those differences correspond to meaningful scene change.

To address this evidence-level ambiguity, we propose \textbf{ChangeJEPA}, which,
to the best of our knowledge, is the first framework to exploit the predictive
latent mechanism of Joint-Embedding Predictive Architectures (JEPA)
\cite{assranSelfSupervisedLearningImages2023a} for remote sensing change
detection. ChangeJEPA reformulates RSCD as a bitemporal post-change latent
completion problem founded on the principle: \emph{predict before you compare}.
JEPA learns semantic structure by predicting the latent representation of a
target from compatible contextual information, without reconstructing the
target in pixel space
\cite{bardes2024revisitingfeaturepredictionlearning,
assran2025vjepa2selfsupervisedvideo}. We extend this principle to bitemporal
scene understanding by predicting the complete post-change latent
representation from the full pre-change image and partial post-change evidence.
Specifically, ChangeJEPA forms two complementary hypotheses of the post-change
scene: a prior prediction conditioned solely on the pre-change image, and an
updated prediction additionally conditioned on visible post-change evidence.
Both hypotheses are trained to match the same complete post-change target
representation and are therefore expressed within a common post-state latent
space. The prior represents the post-state inferred from the historical scene,
whereas the updated prediction incorporates the available post-event context
and moves closer to the observed target wherever that context is sufficient.
Comparisons among the prior, updated, and observed post-change representations
therefore reveal what is expected from the pre-change scene, what can be
explained by the visible post-change evidence, and what remains inconsistent
with both predictions. Rather than requiring the decoder to interpret
discrepancies between independently encoded observations, ChangeJEPA provides a
semantically aligned evidence space in which meaningful scene changes are more
readily distinguished from nuisance-induced appearance variation.

\begin{figure*}[t]
    \centering
    \includegraphics[width=1\linewidth]{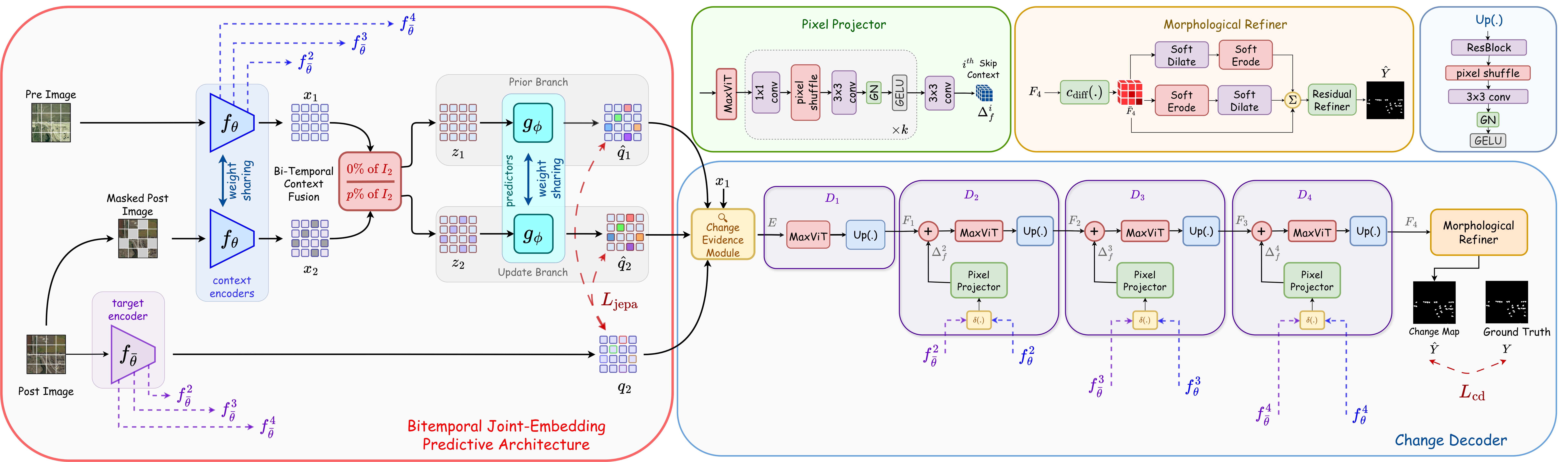}
\caption{\textbf{Overall architecture of ChangeJEPA.}
BiT-JEPA employs two weight-sharing context branches \(f_{\theta}\) and a
target encoder \(f_{\bar{\theta}}\) to represent the complete pre-change image,
the visible post-change patches, and the complete post-change image.
Bitemporal context fusion forms a prior context \(z_1\) with \(0\%\)
post-change visibility and an updated context \(z_2\) with \(p\%\) visible
post-change evidence. Weight-sharing predictors \(g_{\phi}\) map these contexts
to complete post-change predictions \(\widehat{q}_1\) and
\(\widehat{q}_2\), both aligned with the target \(q_2\) through
\(\mathcal{L}_{\mathrm{JEPA}}\). The Change Evidence Module constructs five
complementary comparisons among \(x_1\), \(q_2\), \(\widehat{q}_1\), and
\(\widehat{q}_2\), and aggregates them into \(E\). A four-stage MaxViT decoder
progressively restores spatial resolution while injecting pixel-projected
multilevel bitemporal skip contexts \(\Delta_f^i\) through learnable gates.
Finally, differentiable morphological refinement produces the change map
\(\widehat{Y}\), supervised by \(\mathcal{L}_{\mathrm{cd}}\).}
    \label{fig:oa}
\end{figure*}

{Our main contributions are summarised as follows.} 1) We introduce \textbf{ChangeJEPA}, the first JEPA-based RSCD framework, which reformulates change detection as latent post-state prediction. 2) We propose dual prior and updated post-change predictions, whose complementary discrepancies provide semantically structured evidence for dense change segmentation. 3) Experiments on SYSU-CD, LEVIR-CD+, and CDD demonstrate consistent
improvements over a decoder-matched Siamese baseline, with ChangeJEPA achieving
the highest IoU, F1-score, and overall accuracy on all three benchmarks.



For a comprehensive discussion of related work, we refer the reader to the supplementary material.
\providecommand{\catc}{\mathbin{\vcenter{\hbox{\scriptsize\textcircled{c}}}}}

\section{Methodology}

\subsection{Bitemporal JEPA}
\label{subsec:bitemporal_jepa}

The proposed Bitemporal Joint Embedding Predictive Architecture (BiT-JEPA)
predicts the complete post-change latent representation under two levels of
conditioning. As illustrated in Fig.~\ref{fig:oa}, BiT-JEPA comprises two
weight-sharing context branches realised by $f_{\theta}$, a target encoder
$f_{\bar{\theta}}$, and two weight-sharing predictor branches realised by
$g_{\phi}$. The context branches process the complete pre-change image and the
visible post-change patches, respectively, while the target encoder processes
the complete post-change image to provide the latent prediction target.

Bitemporal context fusion combines the context-encoder outputs to construct an $I_1$ only representation $z_1$, corresponding to $0\%$ post-change conditioning, and a bitemporal representation $z_2$, conditioned on a visible proportion $p\%$ of $I_2$ where $p\in[0.2,0.4]$. The shared predictors then map these representations to the complete post-change predictions $\hat{q}_1$ and $\hat{q}_2$, respectively.

Throughout this section, $I_1,I_2\in\mathbb{R}^{3\times H\times W}$ denote the
pre-change and post-change images, respectively. For an I-JEPA backbone with patch size $P$,
the latent grid has spatial dimensions $h=H/P$ and $w=W/P$, yielding $N=hw$
patch tokens. The embedding dimension is denoted by $C_{\mathrm{emb}}$.

\subsubsection{Targets}
\label{subsubsec:targets}

Following I-JEPA, the post-change image $I_2$ is divided into $N$
non-overlapping patches. A Vision Transformer (ViT) \cite{dosovitskiy2021an} based target encoder $f_{\bar{\theta}}$
processes the complete image and produces the post-change target
representations as, 
\begin{equation}
q_2
=
f_{\bar{\theta}}(I_2)
=
\{q_{2,j}\}_{j=1}^{N},
\label{eq:eq1}
\end{equation}
where $q_2\in\mathbb{R}^{N\times C_{\mathrm{emb}}}$ and
$q_{2,j}\in\mathbb{R}^{C_{\mathrm{emb}}}$ denotes the latent representation of
the $j$-th patch. These representations define the observed post-change latent
state used to supervise the BiT-JEPA predictions.

In addition to the final target representation, intermediate features are
extracted from the eighth, sixteenth, and twenty-fourth ViT blocks of
the target encoder. These features are denoted by
$f_{\bar{\theta}}^{2}$, $f_{\bar{\theta}}^{3}$, and
$f_{\bar{\theta}}^{4}\in\mathbb{R}^{N\times C_{\mathrm{emb}}}$, and are later
used by the change decoder to provide multi-level post-change information for
recovering fine spatial detail.

\subsubsection{Context}
\label{subsubsec:context}

The context is constructed from the complete pre-change image and a visible
subset of the post-change image. To form the post-change context, large
rectangular blocks are randomly masked from the post-change patch grid. These
blocks may overlap, and the hidden-patch ratio is sampled from $[0.6,0.8]$,
corresponding to a visible proportion $p\in[0.2,0.4]$. Let
$\mathcal{V}_2$ denote the set of visible post-change patch indices after
masking. The visible post-change context can be represented as, $x_2
=
f_{\theta}(I_2;\mathcal{V}_2)
=
\{x_{2,j}\}_{j\in\mathcal{V}_2},$
where
$x_2\in\mathbb{R}^{|\mathcal{V}_2|\times C_{\mathrm{emb}}}$ and
$x_{2,j}\in\mathbb{R}^{C_{\mathrm{emb}}}$ denotes the context token
corresponding to the $j$-th visible patch.

Similar to Eq.~\eqref{eq:eq1}, the pre-change image $I_1$ is divided into $N$ non-overlapping patches and
processed by the shared context encoder to obtain the complete pre-change
context as,$
x_1
=
f_{\theta}(I_1)
=
\{x_{1,j}\}_{j=1}^{N},
$, 
where $x_1\in\mathbb{R}^{N\times C_{\mathrm{emb}}}$ and
$x_{1,j}\in\mathbb{R}^{C_{\mathrm{emb}}}$ denotes the pre-change latent
representation of the $j$-th patch.

Intermediate features are also extracted from the eighth, sixteenth, and
twenty-fourth transformer blocks of the context encoder processing $I_1$.
These features are denoted by $f_{\theta}^{2}$, $f_{\theta}^{3}$, and
$f_{\theta}^{4}\in\mathbb{R}^{N\times C_{\mathrm{emb}}}$. They are paired with
the corresponding target-encoder features from $I_2$ to form multi-level,
fine-grained bitemporal skip evidence for the change decoder.

\paragraph{Bitemporal Context Fusion.}
As shown in Fig.~\ref{fig:oa}, the shared context encoder produces the complete
pre-change tokens $x_1$ and the visible post-change tokens $x_2$. The
post-change tokens are first scattered back to their corresponding spatial
locations, while the hidden locations are filled with a learnable mask
embedding $e_m$. The token at the $j$-th location is therefore, $\widetilde{x}_{2,j}
=
v_jx_{2,j}
+
(1-v_j)e_m,$
where $v_j=1$ for a visible post-change patch and $v_j=0$ otherwise. Then we formulate a bitemporal base context as, $b_{2,j}
=
\widetilde{x}_{2,j}
+
(1-v_j)x_{1,j}.$
Thus, visible locations retain the observed post-change tokens, whereas hidden
locations combine the mask embedding with the corresponding pre-change token.

To learn interactions between the two temporal representations, the pre-change
and post-change token grids are processed by a  MaxViT \cite{tu2022maxvitmultiaxisvisiontransformer} based fusion
block $F_{\psi}$. Within $F_{\psi}$, the token grids are reshaped into spatial
feature maps, concatenated channel-wise, processed by the MaxViT
layers, and reshaped back into a token sequence. The updated bitemporal context
is then expressed as, $z_2
=
b_2
+
\gamma F_{\psi}
\left(
x_1,\widetilde{x}_2
\right),$
where $\gamma$ is a learnable scalar initialised to zero, to preserve the
explicit bitemporal context at initialisation while allowing $F_{\psi}$ to
progressively learn temporal interactions.

For the $0\%$-visible prior branch, no observed post-change evidence is
provided, and hence $v_j=0$ at every spatial location. The prior base context
therefore becomes $b_{1,j}=x_{1,j}+e_m$, while the corresponding fused context
is $z_1=b_1+\gamma F_{\psi}(x_1,e_m)$. The prior and updated branches
employ the same fusion architecture, differing only in the availability of
visible post-change evidence.

\begin{figure}
    \centering
    \includegraphics[width=0.95\linewidth]{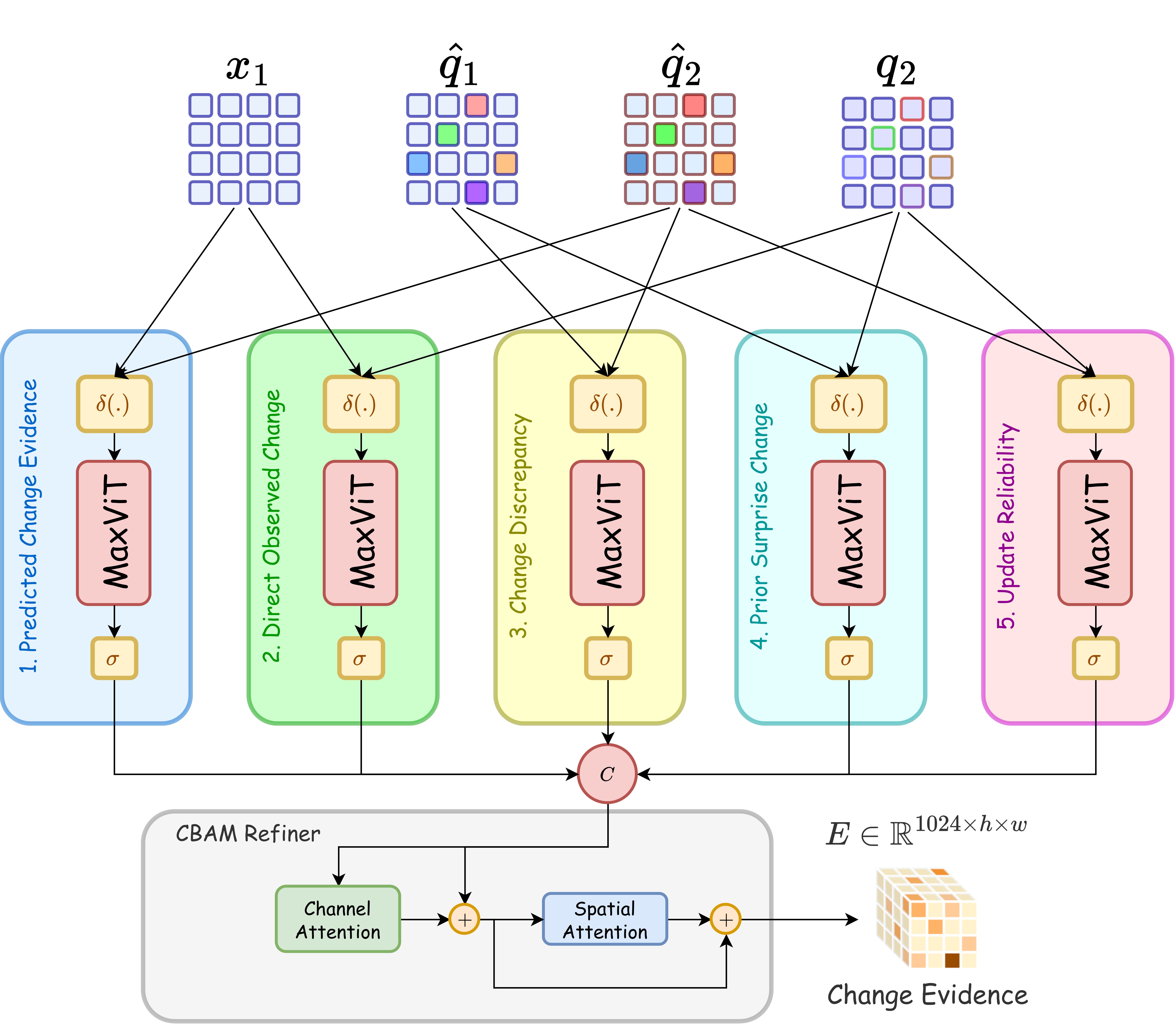}
    \caption{\textbf{Change Evidence Module.}
Five complementary streams are formed from \(x_1\), \(\widehat{q}_1\),
\(\widehat{q}_2\), and \(q_2\), capturing predicted change, direct observed
change, prediction update, prior surprise, and update reliability. Each stream
is encoded by \(\delta(\cdot,\cdot)\), refined independently by MaxViT, and
aggregated through residual channel and spatial attention to produce
\(E\in\mathbb{R}^{1024\times h\times w}\).}
    \label{fig:evidence}
\end{figure}

\subsubsection{Prediction}
\label{subsubsec:prediction}

Given the fused contexts, the predictor estimates the complete post-change
latent representation. We use two weight-sharing predictor branches, a prior
branch operating without observed post-change evidence and an updated branch
conditioned on the visible post-change context. Their predictions are $\widehat{q}_1
=
g_{\phi}(z_1),
\;\;
\widehat{q}_2
=
g_{\phi}(z_2),$
where $\widehat{q}_1$ denotes the prior post-change prediction and
$\widehat{q}_2$ denotes the prediction updated using the visible post-change
evidence.

\begin{figure}[ht]
    \centering
    \includegraphics[width=1\linewidth]{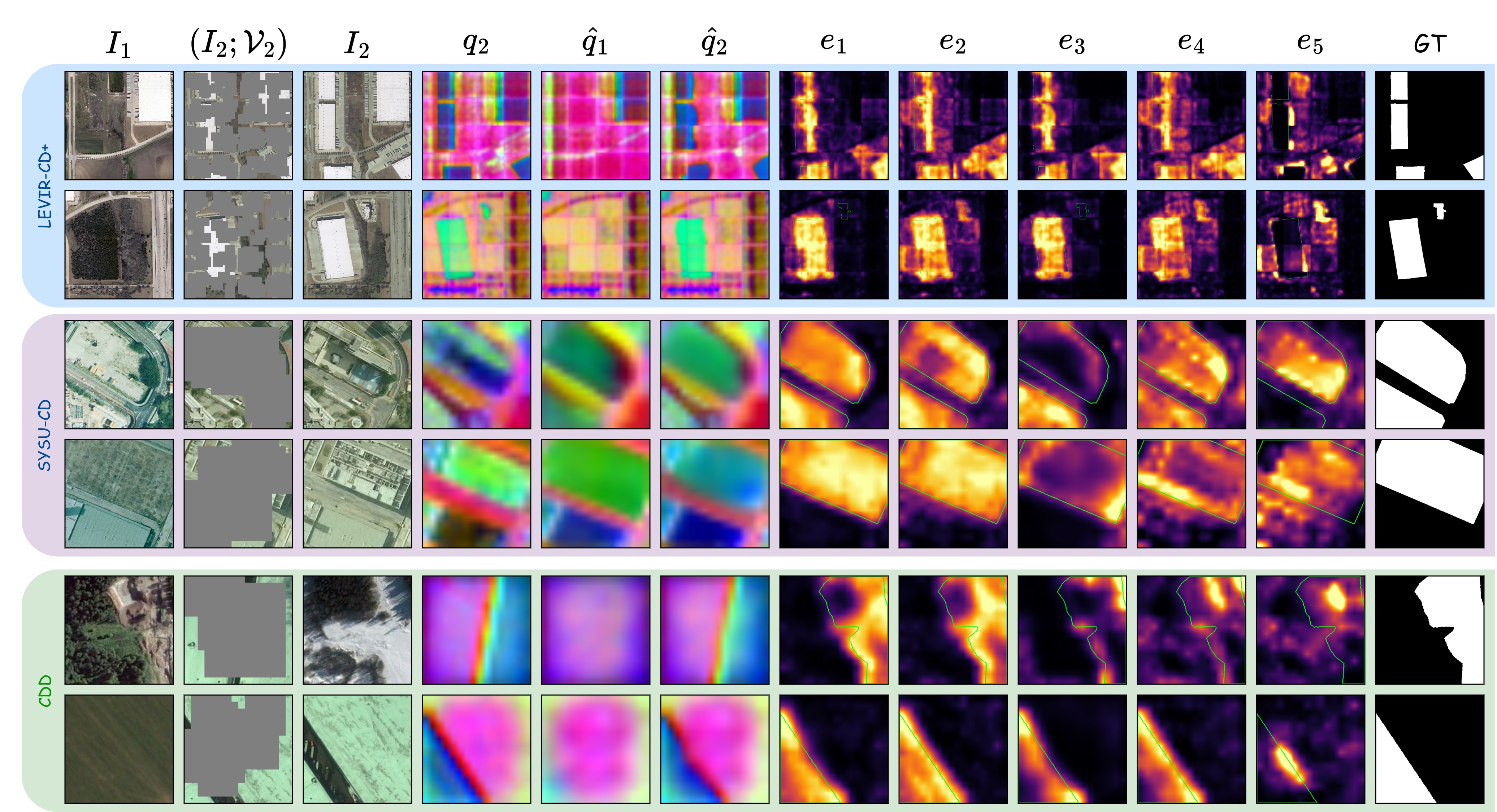}

\caption{\textbf{BiT-JEPA latent predictions and change evidence.}
Shown are the pre-change image \(I_1\), visible post-change context
\((I_2;\mathcal{V}_2)\), complete post-change image \(I_2\), target \(q_2\),
prior prediction \(\widehat{q}_1\), updated prediction \(\widehat{q}_2\), and
the cosine-distance maps of the five evidence branches \(e_1\)--\(e_5\).
Green contours denote the ground-truth change boundaries. The latent representations are visualised using principal component
analysis.}

    \label{fig:targets}
\end{figure}

\subsection{Change Decoder}
\label{subsec:change_decoder}

The change decoder converts the latent states produced by BiT-JEPA into a dense
binary change map. To make their comparisons well defined, the pre-change
feature $x_1$, the observed post-change feature $q_2$, the prior prediction
$\widehat{q}_1$, and the updated prediction $\widehat{q}_2$ are first reshaped
from token sequences into spatial feature maps
$\mathbb{R}^{C_{\mathrm{emb}}\times h\times w}$ and projected into a common
change-detection feature space using a lightweight convolutional projection
neck. The resulting features are aligned on the same spatial grid. For
simplicity, the same notation is retained after reshaping and projection.

\paragraph{Change Evidence Module}Rather than relying on a single bitemporal difference, the Change Evidence
Module constructs five complementary evidence streams, as illustrated in
Fig.~\ref{fig:evidence}. For two feature maps $a$ and $b$, their pairwise
evidence descriptor is defined as $\delta(a,b)
=
a
\catc
|a-b|
\catc
b
\catc
d_{\cos}(a,b),$
where $\catc$ denotes channel-wise concatenation and
$d_{\cos}(a,b)$ denotes the cosine-distance map computed along
the channel dimension. The five
evidence streams are defined as $e_1=\delta(x_1,\widehat{q}_2),
e_2=\delta(x_1,q_2),
e_3=\delta(\widehat{q}_1,\widehat{q}_2),
e_4=\delta(\widehat{q}_1,q_2),
e_5=\delta(\widehat{q}_2,q_2).$
Here, $e_1$ provides \emph{Predicted Change Evidence} by comparing the
pre-change state with the updated post-change prediction. The
\emph{Direct Observed Change} stream $e_2$ captures the conventional Siamese
pre--post discrepancy. More importantly, the prior prediction
$\widehat{q}_1$ is formed from the pre-change scene alone and therefore
represents the post-state expected without any direct clue about what has
actually changed. By contrast, the updated prediction $\widehat{q}_2$
incorporates the visible post-change context and can therefore account for
changed content supported by the later observation. Their comparison in $e_3$
thus measures how much the model must revise its initial expectation after
observing the post-change image, and we term this stream
\emph{Change Discrepancy}. The \emph{Prior Surprise Change} stream $e_4$
measures the departure of the observed post-change state from the prior
prediction, while the \emph{Update Reliability} stream $e_5$ measures the
consistency between the updated prediction and the observed post-change state.

To model each form of change evidence independently, every stream is processed
by a dedicated MaxViT block followed by an activation function, $r_i
=
\sigma\!\left(
\operatorname{maxViT}_i(e_i)
\right),
\;\;
i\in\{1,\ldots,5\}.$
As illustrated in figure \ref{fig:evidence}, the resulting evidence branch features are concatenated and passed through a lightweight
CBAM \cite{woo2018cbamconvolutionalblockattention} based aggregation module to obtain, $E
=
\operatorname{CBAM}
\left(
\text{Concat}[r_1,r_2,r_3,r_4,r_5]
\right),$
where $E\in\mathbb{R}^{1024\times h\times w}$ denotes the aggregated change
evidence.

\begin{figure*}[!t]
    \centering
    \includegraphics[width=1\linewidth]{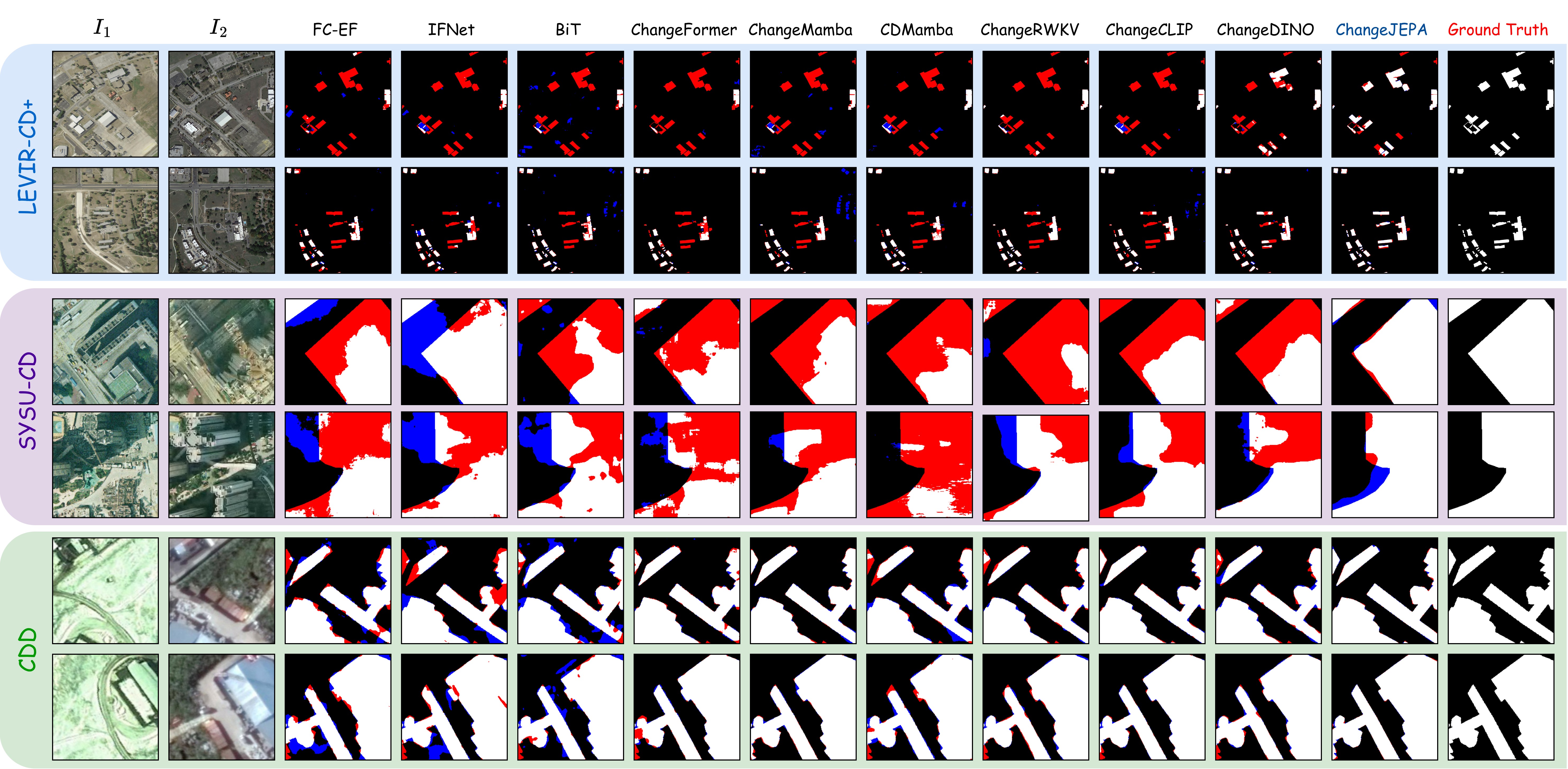}

\caption{\textbf{Qualitative comparison on LEVIR-CD+, SYSU-CD, and CDD.}
White, red, and blue denote true positives, false negatives, and false positives,
respectively. ChangeJEPA yields cleaner, more complete change regions with
better boundary alignment across diverse scene and appearance variations.}

    \label{fig:cdd}
\end{figure*}

\paragraph{4-Stage Decoder.}
The aggregated evidence $E\in\mathbb{R}^{1024\times h\times w}$ provides a
compact semantic description of change at the patch-grid resolution. To recover
the spatial detail required for dense prediction, we progressively decode this
representation through four MaxViT stages, as shown in
Fig.~\ref{fig:oa}. In the first stage, a MaxViT block models local and
long-range interactions within the aggregated evidence. The resulting feature
is then passed through the first upsampling block as given by, $F_1
=
\operatorname{Up}_1
\left(
\operatorname{MaxViT}^{\mathrm{dec}}_1(E)
\right).$
Here, $\operatorname{Up}_1(\cdot)$ performs residual convolutional refinement
followed by learned $2\times$ resolution lifting using channel expansion,
PixelShuffle \cite{shiRealTimeSingleImage2016}, a $3\times3$ convolution, group normalisation (GN) \cite{wuGroupNormalization2018}, and GELU
activation.

Even though $E$ contains strong high-level change evidence, its coarse spatial
resolution is insufficient for accurately recovering small objects and change
boundaries. We therefore introduce multi-level bitemporal skip contexts into
the second, third, and fourth decoder stages. For each stage
$i\in\{2,3,4\}$, the corresponding intermediate pre-change and post-change
features $f_{\theta}^{i}$ and $f_{\bar{\theta}}^{i}$ are first compared using
the pairwise evidence descriptor: $d_f^i
=
\delta
\left(
f_{\theta}^{i},
f_{\bar{\theta}}^{i}
\right),
\;\;
i\in\{2,3,4\}.$
This preserves the bitemporal feature content together with its absolute and
cosine discrepancies, allowing the skip path to provide change-aware spatial
detail rather than single-image features alone.

To model the spatial relationships within each skip descriptor and map it to
the resolution required by the corresponding decoder stage, we process it
through an independent pixel projector: $\Delta_f^i
=
\mathcal{P}_i(d_f^i),
\;\;
i\in\{2,3,4\},$
where $\mathcal{P}_i$ denotes the $i^{th}$ pixel projector and $\Delta_f^i$ is the
resulting skip context. As illustrated in Fig.~\ref{fig:oa}, each pixel
projector first applies a dedicated MaxViT block to model the spatial
relationships within the corresponding bitemporal descriptor. The refined
feature is then passed through $k_i$ successive learned $2\times$ upsampling
steps, where $k_2=1$, $k_3=2$, and $k_4=3$ for the second, third, and fourth
decoder stages, respectively. Each upsampling step comprises a $1\times1$
channel projection, PixelShuffle, a $3\times3$ convolution, group
normalisation, and GELU activation. A final $3\times3$ convolution maps the
feature to the channel width required by the corresponding decoder stage.

To combine the high-level decoder representation with the corresponding
bitemporal spatial context, we project $F_{i-1}$ and $\Delta_f^i$ to the same
channel width and fuse them through gated element-wise addition. The fused
feature is then processed by the corresponding MaxViT and upsampling blocks:
\begin{equation}
F_i
=
\operatorname{Up}_i
\left(
\operatorname{MaxViT}^{\mathrm{dec}}_i
\left(
\mathcal{P}^{F}_i(F_{i-1})
+
\gamma_i\mathcal{P}^{\Delta}_i(\Delta_f^i)
\right)
\right),
\end{equation}
where $\mathcal{P}^{F}_i$, $\mathcal{P}^{\Delta}_i$ denote the corresponding
channel projections, $\gamma_i$ is a learnable skip gate initialised to
zero and $i\in \{2,3,4\}$.

The four stages use progressively reduced channel dimensions of $512$, $256$,
$128$, and $64$, whilst each upsampling block doubles the spatial resolution.
The decoder therefore begins with the aggregated predictive evidence and
progressively introduces finer bitemporal context, allowing semantic change
information to be preserved whilst recovering precise object boundaries and
small changed regions.

\paragraph{Morphological Refiner.}
The final decoder feature $F_4$ is first mapped to a two-channel logit map
$\ell=[\ell_{\mathrm{unchanged}},\ell_{\mathrm{changed}}]$. Since binary change
depends on the relative evidence between these two classes,
$c_{\mathrm{diff}}(\cdot)$ computes their difference and converts it into a
coarse change-probability map as, $p
=
c_{\mathrm{diff}}(\ell)
=   
\sigma\!\left(
\ell_{\mathrm{changed}}
-
\ell_{\mathrm{unchanged}}
\right).$
To suppress isolated responses and fill small holes, we apply differentiable
soft opening and closing \cite{changeDINO,franchiDeepMorphologicalNetworks2020} $p_{\mathrm{open}}
=
\mathcal{D}\!\left(\mathcal{E}(p)\right),
\;\;\;
p_{\mathrm{close}}
=
\mathcal{E}\!\left(\mathcal{D}(p)\right),$
where $\mathcal{E}$ and $\mathcal{D}$ denote soft erosion and dilation,
respectively. The original, opened, and closed responses are then combined and
passed through a lightweight residual refiner to obtain the final change map, $\widehat{Y}\in\mathbb{R}^{2\times H\times W}$.

\subsection{ChangeJEPA Baseline}
\label{subsec:changejepa_baseline}

To determine whether the proposed predictive formulation offers an advantage
over conventional Siamese comparison, we construct a controlled ChangeJEPA
baseline that retains the same I-JEPA initialisation and change decoder. The
complete images $I_1$ and $I_2$ are independently processed by the shared
context encoder and predictor, yielding the projected representations $z_t
=
g_{\phi}\!\left(f_{\theta}(I_t)\right),
\;\;
t\in\{1,2\}.$
The baseline replaces the five predictive evidence streams with a single
Siamese evidence stream as $e_{\mathrm{S}}=\delta(z_1,z_2)$.
The corresponding intermediate encoder features provide direct bitemporal skip
evidence, while the remaining evidence aggregation, MaxViT decoder, and
morphological refiner are unchanged, providing a direct assessment of
predictive change modelling against standard Siamese feature comparison.

\subsection{Loss}
\label{subsec:loss}

ChangeJEPA is trained using predictive latent and supervised change-detection
objectives. Following I-JEPA~\cite{assranSelfSupervisedLearningImages2023a}, the detached and normalised
post-change target and predictive loss are defined as 
\begin{equation}
\mathcal{L}_{\mathrm{JEPA}}
=
\ell\!\left(\widehat{q}_2,\bar{q}_2\right)
+
\lambda_{\mathrm{prior}}
\ell\!\left(\widehat{q}_1,\bar{q}_2\right),
\end{equation}
where $\bar{q}_2
=
\operatorname{LN}\!\left(\operatorname{sg}(q_2)\right),$ and $\ell(\cdot,\cdot)$ denotes the mean token-wise Smooth-$L_1$ distance \cite{7410526}.  The change-detection branch is supervised using the widely adopted weighted
binary cross-entropy, Tversky \cite{salehiTverskyLossFunction2017}, and Lovasz losses \cite{bermanLovaszSoftmaxLossTractable2018} as,
\begin{equation}
\mathcal{L}_{\mathrm{cd}}
=
\mathcal{L}_{\mathrm{BCE}}
+
\mathcal{L}_{\mathrm{Tv}}
+
\lambda_{\mathrm{Lov}}\mathcal{L}_{\mathrm{Lov}}.
\end{equation}
Only $\mathcal{L}_{\mathrm{JEPA}}$ is optimised during predictive warm-up.
Thereafter, the overall objective is $\mathcal{L}
=
\mathcal{L}_{\mathrm{cd}}
+
\lambda_{\mathrm{JEPA}}\mathcal{L}_{\mathrm{JEPA}}.$
The ChangeJEPA baseline contains no predictive objective and is trained solely
on the change-detection dataset using $\mathcal{L}_{\mathrm{cd}}$.   Across all experiments, we set
$\lambda_{\mathrm{prior}}=0.25$,
$\lambda_{\mathrm{JEPA}}=1.0$,  
$\lambda_{\mathrm{Lov}}=0.5$, and
$\alpha=\beta=0.5$ for the Tversky loss. The predictive warm-up lasts for
$30{,}000$ iterations.

\section{Experiments}

\paragraph{Datasets.}
We evaluate ChangeJEPA on three public binary change-detection benchmarks:
LEVIR-CD+ \cite{chenSpatialTemporalAttentionBasedMethod2020},
SYSU-CD \cite{9467555}, and
CDD \cite{lebedevCHANGEDETECTIONREMOTE2018}.
LEVIR-CD+ primarily captures very-high-resolution building changes, SYSU-CD
contains diverse urban and land-cover transitions, and CDD includes pronounced
seasonal appearance variations. We retain the designated training and test splits for all experiments. Complete dataset statistics and split-construction details are provided in the supplementary material.

\noindent{\bf{Evaluation Protocol.}} 
Performance is assessed using Precision, Recall, F1-score, intersection over
union (IoU), and overall accuracy (OA). Formal metric definitions and additional implementation
details are provided in the supplementary material.\\
\noindent{\bf{Implementation Details.}}
All experiments are implemented in PyTorch using $256\times256$ bitemporal
image pairs. The model is optimised using AdamW \cite{loshchilov2018decoupled} with cosine learning-rate
decay \cite{loshchilov2017sgdr}. Standard geometric and photometric augmentations
are applied during training. Detailed training schedules, optimisation
parameters, data preparation procedures, and augmentation settings are
reported in the supplementary material.

\section{Results and Discussion}
\label{sec:results}

\begin{table}[t]
\centering
\caption{Ablation study on SYSU-CD.
$I$, $F_1$, $P$, and $R$ denote IoU, F1-score, Precision, and Recall,
respectively. All values are reported as percentages.}
\label{tab:ablation_studies}
\renewcommand{\arraystretch}{1.08}
\setlength{\tabcolsep}{2.0pt}
\scriptsize

\resizebox{\columnwidth}{!}{%
\begin{tabular}{@{}lccccc|cc|ccccc@{}}
\toprule
&
\multicolumn{5}{c|}{\textbf{Evidence Streams}}
&
\multicolumn{2}{c|}{\textbf{Decoder}}
&
\multicolumn{5}{c}{\textbf{SYSU-CD}} \\
\cmidrule(lr){2-6}
\cmidrule(lr){7-8}
\cmidrule(l){9-13}

\textbf{Variant}
&
$e_1$
&
$e_2$
&
$e_3$
&
$e_4$
&
$e_5$
&
$\Delta_f$
&
\textbf{Morph.}
&
$I$
&
$F_1$
&
\textbf{OA}
&
$P$
&
$R$ \\
\midrule

Siamese baseline
&  & $\checkmark$ &  &  &
& $\checkmark$ & $\checkmark$
& 73.61
& 84.80
& 92.78
& {84.26}
& {85.34}\\

w/o prior prediction    
& $\checkmark$ & $\checkmark$ &  &  & $\checkmark$
& $\checkmark$ & $\checkmark$
& 74.19 & 85.18 & 93.14 & 86.77 & 83.65 \\

w/o updated prediction
&  & $\checkmark$ &  & $\checkmark$ &
& $\checkmark$ & $\checkmark$
& 74.11 & 85.13 & 92.76 & 82.59 & 87.82 \\

w/o direct evidence
& $\checkmark$ &  & $\checkmark$ & $\checkmark$ & $\checkmark$
& $\checkmark$ & $\checkmark$
& 73.76 & 84.90 & 92.86 & 84.67 & 85.14 \\

w/o skip contexts
& $\checkmark$ & $\checkmark$ & $\checkmark$ & $\checkmark$ & $\checkmark$
&  & $\checkmark$
& 73.88 & 84.98 & 92.88 & 84.56 & 85.40 \\

w/o morphological refiner
& $\checkmark$ & $\checkmark$ & $\checkmark$ & $\checkmark$ & $\checkmark$
& $\checkmark$ &
& 74.97 & 85.69 & 93.22 & 85.33  & 86.06 \\

\cj{\textbf{ChangeJEPA}}
& \cj{$\checkmark$} & \cj{$\checkmark$} & \cj{$\checkmark$}
& \cj{$\checkmark$} & \cj{$\checkmark$}
& \cj{$\checkmark$} & \cj{$\checkmark$}
& \cj{\textbf{75.12}} & \cj{\textbf{85.79}}
& \cj{\textbf{93.34}} & \cj{\textbf{86.37}} & \cj{\textbf{85.22}} \\

\bottomrule

\end{tabular}%
}

\parbox{\columnwidth}{\scriptsize
The Siamese baseline uses
$e_{\mathrm{S}}=\delta(z_1,z_2)$ instead of the predictive evidence streams.
Here, $\Delta_f$ denotes the multi-level skip contexts
$\{\Delta_f^i\}_{i=2}^{4}$; an empty entry indicates that the corresponding
component is omitted.}
\end{table}

\begin{table*}[t]
\centering
\caption{Quantitative comparison on SYSU-CD, LEVIR-CD+, and CDD.
For each dataset, all models are trained and evaluated using the same
train--test split. Pre and Rec denote Precision and Recall, respectively.
All values are reported as percentages. The best and second-best results
for each metric are highlighted using \best{boldface} and
\second{underlining}, respectively.}
\label{tab:quantitative_comparison}
\setlength{\tabcolsep}{2.5pt}
\scriptsize
\begin{tabular}{
@{}
l
@{\hspace{12pt}}
*{5}{c}
@{\hspace{8pt}}
*{5}{c}
@{\hspace{8pt}}
*{5}{c}
@{}
}
\toprule
\multirow{2}{*}{\textbf{Method}}
& \multicolumn{5}{c}{\textbf{SYSU-CD}}
& \multicolumn{5}{c}{\textbf{LEVIR-CD+}}
& \multicolumn{5}{c}{\textbf{CDD}} \\
\cmidrule(lr){2-6}
\cmidrule(lr){7-11}
\cmidrule(lr){12-16}
& \textbf{IoU} & \textbf{F1} & \textbf{OA} & \textbf{Pre} & \textbf{Rec}
& \textbf{IoU} & \textbf{F1} & \textbf{OA} & \textbf{Pre} & \textbf{Rec}
& \textbf{IoU} & \textbf{F1} & \textbf{OA} & \textbf{Pre} & \textbf{Rec} \\
\midrule

FC-EF \cite{8451652}
& 64.35 & 78.31 & 89.58 & 76.91 & 79.77
& 58.69 & 73.97 & 97.98 & 78.01 & 70.32
& 85.26 & 92.04 & 98.06 & 93.17 & 90.94 \\

FC-Siam-Diff  \cite{8451652}
& 65.05 & 78.83 & 90.33 & 81.53 & 76.30
& 60.15 & 75.12 & 98.09 & 79.83 & 70.93
& 82.97 & 90.69 & 97.77 & 93.41 & 88.12 \\

FC-Siam-Conc  \cite{8451652}
& 63.85 & 77.94 & 89.67 & 78.52 & 77.36
& 62.28 & 76.75 & 98.12 & 77.37 & 76.15
& 84.70 & 91.72 & 98.01 & 94.31 & 89.26 \\

IFNet \cite{zhangDeeplySupervisedImage2020}
& 67.42 & 80.54 & 90.84 & 80.67 & 80.41
& 68.18 & 81.08 & 98.47 & 81.76 & 80.41
& 87.35 & 93.25 & 98.36 & 94.74 & 91.80 \\

SNUNet \cite{9355573}
& 53.32 & 69.56 & 88.27 & \best{89.70} & 56.80
& 62.78 & 77.14 & 98.18 & 79.14 & 75.24
& 67.04 & 80.27 & 95.50 & 87.59 & 74.07 \\

BIT \cite{BIT}
& 64.79 & 78.63 & 89.82 & 77.82 & 79.46
& 66.43 & 79.83 & 98.31 & 77.89 & 81.87
& 87.91 & 93.57 & 98.44 & 95.10 & 92.08 \\

ChangeFormer \cite{changeformer}
& 66.03 & 79.54 & 90.60 & 81.74 & 77.46
& 64.74 & 78.60 & 98.25 & 78.32 & 78.87
& 94.34 & 97.09 & 99.28 & 97.48 & 96.70 \\

ChangeMamba \cite{changemamba}
& 71.67 & 83.50 & 92.63 & \second{88.45} & 79.07
& 69.63 & 82.09 & 98.50 & 80.17 & 84.12
& \second{95.82} & \second{97.87} & \second{99.47}
& \second{97.85} & 97.88 \\

CDMamba \cite{10902569}
& 68.27 & 81.14 & 91.38 & 83.79 & 78.65
& 67.65 & 80.71 & 98.43 & 80.86 & 80.56
& 89.82 & 94.64 & 98.68 & 94.90 & 94.37 \\

ChangeRWKV \cite{yangQuadraticLinearTimeChange2026}
& 69.61 & 82.08 & 92.02 & 87.30 & 77.45
& 74.96 & 85.69 & 98.84 & \best{86.50} & 84.90
& 92.80 & 96.27 & 99.08 & 96.52 & 96.02 \\

ChangeCLIP \cite{changeclip}
& 72.40 & 83.99 & 92.64 & 86.18 & 81.91
& 73.60 & 84.79 & 98.76 & 84.73 & 84.86
& 95.64 & 97.77 & 99.45 & \best{97.86} & 97.69 \\

ChangeDINO \cite{changeDINO}
& \second{73.95} & \second{85.04} & 92.61
& 81.35 & \best{89.07}
& \second{75.59} & \second{86.10} & \second{98.85}
& 84.82 & \second{87.41}
& 92.59 & 96.15 & 99.05 & 96.41 & 95.90 \\

\cjb{ChangeJEPA-Baseline}
& \cjb{73.61}
& \cjb{84.80}
& \cjb{\second{92.78}}
& \cjb{84.26}
& \cjb{\second{85.34}}
& \cjb{74.89}
& \cjb{85.64}
& \cjb{98.81}
& \cjb{83.90}
& \cjb{\best{87.47}}
& \cjb{95.48}
& \cjb{97.69}
& \cjb{99.43}
& \cjb{97.07}
& \cjb{\second{98.32}} \\

\cj{ChangeJEPA}
& \cj{\best{75.12}}
& \cj{\best{85.79}}
& \cj{\best{93.34}}
& \cj{86.37}
& \cj{85.22}
& \cj{\best{76.05}}
& \cj{\best{86.39}}
& \cj{\best{98.88}}
& \cj{\second{85.90}}
& \cj{86.89}
& \cj{\best{96.56}}
& \cj{\best{98.25}}
& \cj{\best{99.57}}
& \cj{97.76}
& \cj{\best{98.75}} \\

\bottomrule
\end{tabular}%
\end{table*}

\noindent{\bf{Quantitative Analysis.}}
As can be seen from Table~\ref{tab:quantitative_comparison}, ChangeJEPA achieves the
strongest IoU, F1-score, and OA across all evaluated benchmarks. This consistent
ranking is notable because the datasets represent different change-detection
conditions, including heterogeneous land-cover transitions, building-centred
changes, and pronounced seasonal appearance variation. The gains in IoU and
F1-score are particularly meaningful because these metrics directly reflect
change-region quality and are less influenced by the dominant unchanged class
than OA.

Relative to ChangeJEPA-Baseline, the proposed model consistently improves
region overlap while exhibiting dataset-dependent precision--recall behaviour.
On SYSU-CD and LEVIR-CD+, the improvement is driven mainly by greater precision
while maintaining competitive recall, indicating stronger suppression of false
detections. On CDD, both precision and recall improve despite substantial
seasonal variation. These results suggest that predictive conditioning does not
simply strengthen every discrepancy response; it helps the model retain
differences that are more likely to represent meaningful scene change.

\noindent{\bf{Qualitative Analysis.}}
Figure~\ref{fig:cdd} presents representative change maps, where white, red,
blue, and black denote true positives, false negatives, false positives, and
true negatives, respectively. On LEVIR-CD+, ChangeJEPA recovers compact and
disconnected building changes while suppressing responses from neighbouring
roofs and roads. On SYSU-CD, it better preserves narrow and concave unchanged
regions within large change areas, reducing boundary displacement and
over-segmentation. On CDD, it retains thin structures and irregular contours
despite strong seasonal and radiometric differences. In these examples, the
remaining errors are largely confined to ambiguous boundaries rather than
distributed across unchanged regions, demonstrating cleaner and more selective
change localisation.

Figure~\ref{fig:targets} illustrates the mechanism underlying this behaviour.
The prior prediction \(\widehat{q}_1\) represents the post-state inferred from
\(I_1\) alone, whereas the updated prediction \(\widehat{q}_2\) revises this
expectation using visible post-change context. The corresponding cosine maps
exhibit distinct spatial responses: the observed bitemporal relation identifies
where the scene differs, the prior--target relation expresses departure from the
expected post-state, the prior--updated relation captures the revision induced
by new context, and the updated--target relation reveals what remains
unexplained after that revision. These complementary responses support the
intended decomposition of change evidence into observed difference, prior
expectation, contextual explanation, and residual inconsistency. Additional
qualitative examples are reported in the supplementary material.

\subsection{Ablation Studies}
\label{subsec:ablation_studies}

Table~\ref{tab:ablation_studies} demonstrates that the prior and updated
predictions contribute complementary information. Removing the
updated-prediction relations shifts the model towards higher recall but lower
precision, producing broader and less selective detections. This indicates that
visible post-change context is important for explaining ambiguous discrepancies
and suppressing false alarms. Removing the prior-prediction relations produces
the opposite tendency, with reduced recall showing that the
pre-change-conditioned expectation helps expose changes that are less evident
from the updated hypothesis alone. Both removals reduce the aggregate
segmentation quality, confirming that the two predictions are not
interchangeable.

Direct bitemporal evidence also remains essential and causes the strongest
degradation among the evidence ablations when removed. Predictive modelling
therefore complements rather than replaces direct comparison: the observed
relation identifies where the images differ, while the predictive relations
help determine whether those differences are expected, contextually explained,
or unresolved. The multi-level skip contexts further improve spatial
localisation by restoring finer bitemporal structure, while the morphological
refiner provides a smaller but consistent improvement by reducing spurious
responses and boundary leakage.

Finally, ChangeJEPA outperforms its Siamese baseline when both are paired with
the same downstream decoding and refinement design. Together with the component
ablations and qualitative evidence, this result supports the central principle
of ChangeJEPA: meaningful change is represented not only by what differs between
two observations, but also by what remains unexpected after predicting and
updating the post-state.
\section{Conclusion}
\label{sec:conclusion}

We introduced ChangeJEPA, a bitemporal JEPA framework combining direct
comparison with latent post-state prediction. It forms a prior prediction
from the pre-change image and an updated prediction using partial
post-change context. Comparisons with the observed representations
provide complementary evidence of scene differences, contextual revision,
and residual inconsistency. On SYSU-CD, LEVIR-CD+, and CDD, ChangeJEPA outperforms its decoder-matched
Siamese baseline and achieves the highest IoU, F1-score, and overall
accuracy among the evaluated methods. Ablations support distinct
contributions from both predictions whilst establishing the importance
of direct evidence. Qualitative comparisons and perturbation experiments
also support improved localisation under the appearance variations examined. These findings suggest that post-state prediction assists the
interpretation of bitemporal differences. Change may appear in revisions
supported by the later observation or in discrepancies that remain
unexplained. ChangeJEPA thus relates observed differences to predicted
scene structure. Further work will consider multiple observations,
different sensors, and imperfectly registered imagery.




\bibliography{aaai2027}

\twocolumn[{%
  \begin{center}
    {\Large\bfseries Supplementary Material for\par}
    \vspace{0.5em}
    {\Large\bfseries
      ChangeJEPA: Predict Before You Compare\\
      for Remote Sensing Change Detection\par}
    \vspace{1.5em}
  \end{center}
}]

\setcounter{secnumdepth}{1}
\setcounter{section}{0}
\section{Overview}
\label{sec:supp_overview}

This supplement provides further details of the method and its
experimental assessment. It discusses related work on foundation
models for change detection and joint embedding predictive
architectures, and presents additional ablation studies and
qualitative comparisons. The Siamese baseline is described,
together with the datasets, experimental partitions, evaluation
measures, and procedures for training and inference. The final
sections report computational requirements and examine robustness
to controlled variations in illumination, sensor degradation,
and occlusion. The notation is consistent with that used in the
main paper.

\section{Related Work}
\label{sec:related_work}

\noindent\textbf{Foundation-Model-Based Change Detection.}
Recent work has increasingly transferred large-scale pretrained representations
to remote sensing change detection (RSCD). A prominent direction adapts
segmentation foundation models, particularly the Segment Anything Model (SAM)
\cite{kirillovSegmentAnything2023}. SAM-CD combines a frozen FastSAM encoder
with trainable convolutional adaptors and a task-agnostic semantic branch to
recover change-sensitive multiscale features \cite{10443350}. AnyChange instead
performs training-free bitemporal latent matching in SAM's feature space,
using intra-image and inter-image semantic similarities for zero-shot change
detection \cite{zhengSegmentAnyChange2024}. TTP introduces low-rank trainable
parameters into a SAM backbone and employs a time-travelling activation gate
to inject cross-temporal information into the pixel-semantic feature space
\cite{10640593}. Meta-CD similarly couples FastSAM with a lightweight adaptor
to support rapid adaptation under limited supervision \cite{10902491}.

Other approaches exploit vision--language, self-supervised, multimodal, or
generative foundation models. ChangeCLIP adapts CLIP representations for
bitemporal feature extraction and vision--language-guided decoding
\cite{changeclip}. BAN connects a frozen foundation model to a task-specific
bitemporal network through learnable bridging modules \cite{10438490}, whereas
ChangeDINO combines frozen DINOv3 representations with a lightweight trainable
backbone and a differential transformer decoder \cite{changeDINO}. UniChange
integrates a multimodal large language model with a change-perception vision
module to support binary and semantic change detection
\cite{zhangUniChangeUnifyingChange2026}. Changen2 follows a generative
direction, synthesising multitemporal imagery and corresponding change labels
for downstream model training \cite{10713915}.

These methods improve representation transfer, parameter-efficient adaptation,
or the availability of change supervision. Their discriminative inference
paths, however, still construct change evidence primarily by matching,
differencing, or fusing representations extracted from the observed temporal
inputs. Changen2 modifies the training distribution rather than the downstream
inference mechanism. ChangeJEPA addresses this evidence-level limitation
directly: instead of relying solely on discrepancies between observed
bitemporal representations, it first predicts the post-change latent state and
then constructs the evidence used for dense change segmentation.

\begin{figure*}[t]
    \centering
    \includegraphics[width=\textwidth]{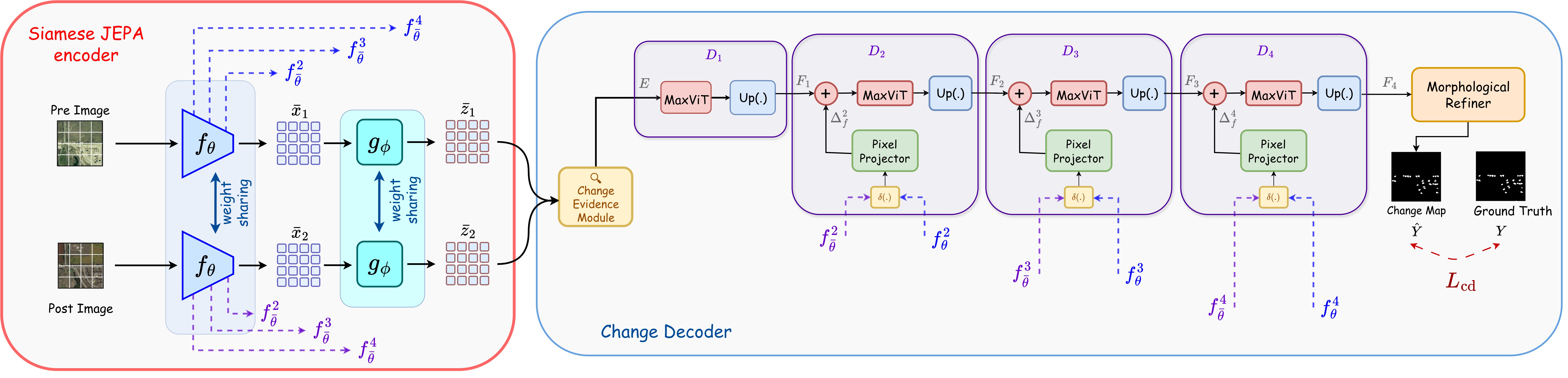}
    \caption{Architecture of the decoder-matched ChangeJEPA baseline. The
    complete pre-change and post-change images are processed independently by
    weight-sharing, I-JEPA-initialised context-encoder and predictor branches.
    A single Siamese evidence stream,
    \(e_{\mathrm{S}}=\delta(z_1,z_2)\), together with direct bitemporal skip
    evidence, replaces the predictive evidence used by ChangeJEPA. The evidence
    aggregation interface, four-stage MaxViT decoder, and morphological refiner
    remain unchanged. The baseline is trained solely with the change-detection
    objective \(\mathcal{L}_{\mathrm{cd}}\).}
    \label{fig:changejepa_baseline}
\end{figure*}

\noindent\textbf{Joint-Embedding Predictive Architectures.}
Joint-Embedding Predictive Architectures (JEPAs) learn by predicting the latent
representation of a target from compatible contextual information rather than
reconstructing the target in observation space
\cite{lecunPathAutonomousMachinea,assranSelfSupervisedLearningImages2023a}.
I-JEPA instantiates this principle for images using a context encoder, a
momentum-updated target encoder, and a predictor that estimates masked
target-block representations from visible context
\cite{assranSelfSupervisedLearningImages2023a}. Subsequent work extends latent
prediction to transformation-conditioned representation learning and
spatio-temporal video modelling
\cite{garrido2024learningleveragingworldmodels,
bardes2024revisitingfeaturepredictionlearning,
assran2025vjepa2selfsupervisedvideo}.

Predictive latent learning has also emerged in remote sensing. SAR-JEPA
predicts multiscale gradient representations for self-supervised SAR automatic
target recognition \cite{liPredictingGradientBetter2024}. AnySat employs a
scale-adaptive JEPA objective to pretrain a general-purpose multimodal
Earth-observation backbone that is subsequently transferred to downstream
tasks \cite{Astruc_2025_CVPR}. REJEPA applies
context-to-target prediction to remote sensing image retrieval
\cite{choudhuryREJEPANovelJointEmbedding2025}, while LEPA predicts
transformation-conditioned satellite-image embeddings to learn geometric
equivariance \cite{scheurer2026lepalearninggeometricequivariance}. T-SAR-JEPA
retains latent prediction at inference, but forecasts representations from
temporal SAR stacks and uses the resulting prediction error for anomaly
detection \cite{woldesenbet2026tsarjepaselfsupervisedtemporalanomaly}.

Nevertheless, existing remote sensing applications employ JEPA principally as a pretraining objective or as a predictor for target recognition, retrieval, geometric adaptation, and temporal anomaly detection. They do not use latent prediction as the decision mechanism for dense bitemporal change segmentation. This leaves unexplored a fundamentally different formulation of RSCD: predicting the complete post-change latent scene from the pre-change observation and partial post-change evidence, and then identifying change from what remains unexplained by that prediction. This gap motivates our \emph{ChangeJEPA} formulation.

\begin{table}[t]
\centering
\caption{Ablation study on SYSU-CD and LEVIR-CD+.
$I$ and $F_1$ denote IoU and F1-score, respectively. All values are reported as percentages.}
\label{tab:ablation_combined}
\renewcommand{\arraystretch}{1.08}
\setlength{\tabcolsep}{2.0pt}
\scriptsize

\resizebox{\columnwidth}{!}{%
\begin{tabular}{@{}lccccc|cc|cc|cc|cc@{}}
\toprule
&
\multicolumn{5}{c|}{\textbf{Evidence Streams}}
&
\multicolumn{2}{c|}{\textbf{Decoder}}
&
\multicolumn{2}{c|}{\textbf{Training}}
&
\multicolumn{2}{c|}{\textbf{SYSU}}
&
\multicolumn{2}{c}{\textbf{LEVIR}} \\
\cmidrule(lr){2-6}
\cmidrule(lr){7-8}
\cmidrule(lr){9-10}
\cmidrule(lr){11-12}
\cmidrule(l){13-14}

\textbf{Variant}
& $e_1$ & $e_2$ & $e_3$ & $e_4$ & $e_5$
& $\Delta_f$ & \textbf{Morph.}
& $\mathcal{L}_{\mathrm{JEPA}}$ & \textbf{WU}
& $I$ & $F_1$
& $I$ & $F_1$ \\
\midrule

Siamese baseline
&  & $\checkmark$ &  &  &
& $\checkmark$ & $\checkmark$
&  &
& 73.61 & 84.80
& 74.89 & 85.64 \\

w/o prior prediction
& $\checkmark$ & $\checkmark$ &  &  & $\checkmark$
& $\checkmark$ & $\checkmark$
& $\checkmark$ & $\checkmark$
& 74.19 & 85.18
& 74.64 & 85.48 \\

w/o updated prediction
&  & $\checkmark$ &  & $\checkmark$ &
& $\checkmark$ & $\checkmark$
& $\checkmark$ & $\checkmark$
& 74.11 & 85.13
& 73.69 & 84.85 \\

w/o direct evidence
& $\checkmark$ &  & $\checkmark$ & $\checkmark$ & $\checkmark$
& $\checkmark$ & $\checkmark$
& $\checkmark$ & $\checkmark$
& 73.76 & 84.90
& 74.73 & 85.54 \\

w/o skip contexts
& $\checkmark$ & $\checkmark$ & $\checkmark$ & $\checkmark$ & $\checkmark$
&  & $\checkmark$
& $\checkmark$ & $\checkmark$
& 73.88 & 84.98
& 75.66 & 86.14 \\

w/o morphological refiner
& $\checkmark$ & $\checkmark$ & $\checkmark$ & $\checkmark$ & $\checkmark$
& $\checkmark$ &
& $\checkmark$ & $\checkmark$
& 74.97 & 85.69
& 75.17 & 85.82 \\

\midrule
\multicolumn{14}{@{}l}{\textit{Stream subsets, capacity, and training}} \\

Direct stream only
&  & $\checkmark$ &  &  &
& $\checkmark$ & $\checkmark$
& $\checkmark$ & $\checkmark$
& 73.67 & 84.84
& 74.91 & 85.66 \\

Prior surprise only
&  &  &  & $\checkmark$ &
& $\checkmark$ & $\checkmark$
& $\checkmark$ & $\checkmark$
& 73.71 & 84.87
& 74.12 & 85.14 \\

$e_1{+}e_2$
& $\checkmark$ & $\checkmark$ &  &  &
& $\checkmark$ & $\checkmark$
& $\checkmark$ & $\checkmark$
& 73.86 & 84.96
& 74.97 & 85.69 \\

Five direct streams ($e_2{\times}5$)
&  & $\checkmark$\rlap{$^{\times5}$} &  &  &
& $\checkmark$ & $\checkmark$
& $\checkmark$ & $\checkmark$
& 73.71 & 84.87
& 74.82 & 85.60 \\

w/o predictive warm-up
& $\checkmark$ & $\checkmark$ & $\checkmark$ & $\checkmark$ & $\checkmark$
& $\checkmark$ & $\checkmark$
& $\checkmark$ &
& 73.21 & 84.53
& 73.80 & 84.93 \\

w/o latent loss$^{*}$
& $\checkmark$ & $\checkmark$ & $\checkmark$ & $\checkmark$ & $\checkmark$
& $\checkmark$ & $\checkmark$
&  &
& \multicolumn{2}{c|}{collapses}
& \multicolumn{2}{c}{collapses} \\

\midrule

\cj{\textbf{ChangeJEPA}}
& \cj{$\checkmark$} & \cj{$\checkmark$} & \cj{$\checkmark$}
& \cj{$\checkmark$} & \cj{$\checkmark$}
& \cj{$\checkmark$} & \cj{$\checkmark$}
& \cj{$\checkmark$} & \cj{$\checkmark$}
& \cj{\textbf{75.12}} & \cj{\textbf{85.79}}
& \cj{\textbf{76.05}} & \cj{\textbf{86.39}} \\

\bottomrule
\end{tabular}%
}

\parbox{\columnwidth}{\scriptsize
All variants use the best checkpoint with test-time augmentation, as in Table~2.
The Siamese baseline uses $e_{\mathrm{S}}=\delta(z_1,z_2)$ instead of the predictive evidence
streams and is trained with $\mathcal{L}_{\mathrm{cd}}$ only.
$\Delta_f$ denotes the multi-level skip contexts $\{\Delta_f^i\}_{i=2}^{4}$;
$\mathcal{L}_{\mathrm{JEPA}}$ marks the predictive loss and WU the 30k-iteration predictive warm-up;
an empty entry indicates that the corresponding component is omitted.
$e_2{\times}5$ replaces all five streams with five independent direct streams through the same fusion.
$^{*}$Training collapses even with a longer decoder warm-up.}
\end{table}


\begin{table}[t]
\centering
\caption{Robustness to misregistration. The post-event image $I_2$ is translated by
2, 4, or 8 pixels (down-right, replicate border) while $I_1$ and the ground truth are
unchanged. Final checkpoints, single forward pass without test-time augmentation.
$I$ and $F_1$ denote IoU and F1-score; all values are reported as percentages.}
\label{tab:misregistration}
\setlength{\tabcolsep}{3.0pt}
\renewcommand{\arraystretch}{1.08}
\scriptsize
\resizebox{\columnwidth}{!}{%
\begin{tabular}{@{}llcccccc@{}}
\toprule
\multirow{2}{*}{\textbf{Dataset}} & \multirow{2}{*}{\textbf{Method}}
& \multicolumn{2}{c}{\textbf{2 px}}
& \multicolumn{2}{c}{\textbf{4 px}}
& \multicolumn{2}{c}{\textbf{8 px}} \\
\cmidrule(lr){3-4}
\cmidrule(lr){5-6}
\cmidrule(l){7-8}
& & $I$ & $F_1$ & $I$ & $F_1$ & $I$ & $F_1$ \\
\midrule

\multirow{2}{*}{LEVIR-CD+}
& \cjb{ChangeJEPA-Baseline}
& \cjb{70.00} & \cjb{82.35} & \cjb{61.65} & \cjb{76.27} & \cjb{47.30} & \cjb{64.22} \\
& \cj{ChangeJEPA}
& \cj{\textbf{71.37}} & \cj{\textbf{83.29}} & \cj{\textbf{62.90}} & \cj{\textbf{77.23}} & \cj{\textbf{48.50}} & \cj{\textbf{65.32}} \\
\midrule

\multirow{2}{*}{SYSU-CD}
& \cjb{ChangeJEPA-Baseline}
& \cjb{72.02} & \cjb{83.73} & \cjb{71.63} & \cjb{83.47} & \cjb{69.51} & \cjb{82.01} \\
& \cj{ChangeJEPA}
& \cj{\textbf{74.34}} & \cj{\textbf{85.28}} & \cj{\textbf{73.51}} & \cj{\textbf{84.73}} & \cj{\textbf{69.59}} & \cj{\textbf{82.07}} \\

\bottomrule
\end{tabular}%
}
\end{table}


\begin{table}[t]
\centering
\caption{Effect of test-time augmentation (TTA). With TTA, ChangeJEPA averages five
fixed-seed block masks over three flips and the baseline averages three flips
(Table~2); without TTA, every method uses a single forward pass (ChangeJEPA: one
fixed-seed mask). $I$ and $F_1$ denote IoU and F1-score; all values
are reported as percentages.}
\label{tab:tta}
\setlength{\tabcolsep}{3.0pt}
\renewcommand{\arraystretch}{1.08}
\scriptsize
\resizebox{\columnwidth}{!}{%
\begin{tabular}{@{}l|cc|cc|cc@{}}
\toprule
\multirow{2}{*}{\textbf{Method}}
& \multicolumn{2}{c|}{\textbf{SYSU-CD}}
& \multicolumn{2}{c|}{\textbf{LEVIR-CD+}}
& \multicolumn{2}{c}{\textbf{CDD}} \\
\cmidrule(lr){2-3}
\cmidrule(lr){4-5}
\cmidrule(l){6-7}
& $I$ & $F_1$ & $I$ & $F_1$ & $I$ & $F_1$ \\
\midrule
\multicolumn{7}{@{}l}{\textit{Single pass, no TTA}} \\
Best competitor$^{\dagger}$
& 73.95 & 85.04 & 75.59 & 86.10 & 95.82 & 97.87 \\
\cjb{ChangeJEPA-Baseline}
& \cjb{72.97} & \cjb{84.37} & \cjb{74.51} & \cjb{85.39} & \cjb{95.23} & \cjb{97.56} \\
\cj{ChangeJEPA}
& \cj{\textbf{74.66}} & \cj{\textbf{85.49}} & \cj{\textbf{75.70}} & \cj{\textbf{86.17}} & \cj{\textbf{96.41}} & \cj{\textbf{98.17}} \\
\midrule
\multicolumn{7}{@{}l}{\textit{With TTA (Table~2)}} \\
\cjb{ChangeJEPA-Baseline}
& \cjb{73.61} & \cjb{84.80} & \cjb{74.89} & \cjb{85.64} & \cjb{95.48} & \cjb{97.69} \\
\cj{ChangeJEPA}
& \cj{\textbf{75.12}} & \cj{\textbf{85.79}} & \cj{\textbf{76.05}} & \cj{\textbf{86.39}} & \cj{\textbf{96.56}} & \cj{\textbf{98.25}} \\
\bottomrule
\end{tabular}%
}

\parbox{\columnwidth}{\scriptsize
$^{\dagger}$Strongest competitor in Table~2 per dataset, evaluated with a single pass:
ChangeDINO on SYSU-CD and LEVIR-CD+, ChangeMamba on CDD.}
\end{table}


\begin{table}[t]
\centering
\caption{Sensitivity to the test-time mask, single forward pass of the final checkpoint. The default
mask hides 70\% of $I_2$ with a fixed seed. Top: the seed is varied at 70\% hidden; bottom: the hidden
ratio is varied with the seed fixed. $\Delta I$ is the IoU change in percentage points relative to the
default mask.}
\label{tab:mask_sensitivity}
\setlength{\tabcolsep}{6pt}
\renewcommand{\arraystretch}{1.08}
\scriptsize
\begin{tabular}{@{}lcc@{}}
\toprule
& \textbf{SYSU-CD} & \textbf{LEVIR-CD+} \\
\midrule
\multicolumn{3}{@{}l}{\textit{Mask seed varied (10 seeds, 70\% hidden)}} \\
\quad Std.\ of IoU                  & 0.003 & 0.002 \\
\quad Largest $|\Delta I|$          & 0.005 & 0.003 \\
\midrule
\multicolumn{3}{@{}l}{\textit{Hidden ratio varied (seed fixed): $\Delta I$ vs.\ 70\%}} \\
\quad 50\% hidden                   & $+$0.003 & $+$0.013 \\
\quad 60\% hidden                   & $+$0.004 & $+$0.006 \\
\quad 80\% hidden                   & $-$0.002 & $-$0.005 \\
\quad 90\% hidden                   & $-$0.009 & $-$0.009 \\
\quad Largest $|\Delta I|$          & 0.009 & 0.013 \\
\bottomrule
\end{tabular}
\end{table}

\begin{table}[t]
\centering
\caption{Hidden-token filling on SYSU-CD. Hidden post-event locations are filled with the mask
embedding plus the same-index pre-event token (ours) or with the mask embedding only; both models
are trained identically and evaluated at their best checkpoint with test-time augmentation.
Entries are changes relative to ours.}
\label{tab:mask_filling}
\setlength{\tabcolsep}{6pt}
\renewcommand{\arraystretch}{1.08}
\scriptsize
\begin{tabular}{@{}lcc@{}}
\toprule
\textbf{Hidden-token filling} & $\Delta I$ & $\Delta F_1$ \\
\midrule
Mask embedding $+$ same-index $I_1$ token (ours) & 0 & 0 \\
Mask embedding only                              & $-$0.54 & $-$0.36 \\
\bottomrule
\end{tabular}
\end{table}

\begin{figure*}
    \centering
    \includegraphics[width=1.05\linewidth]{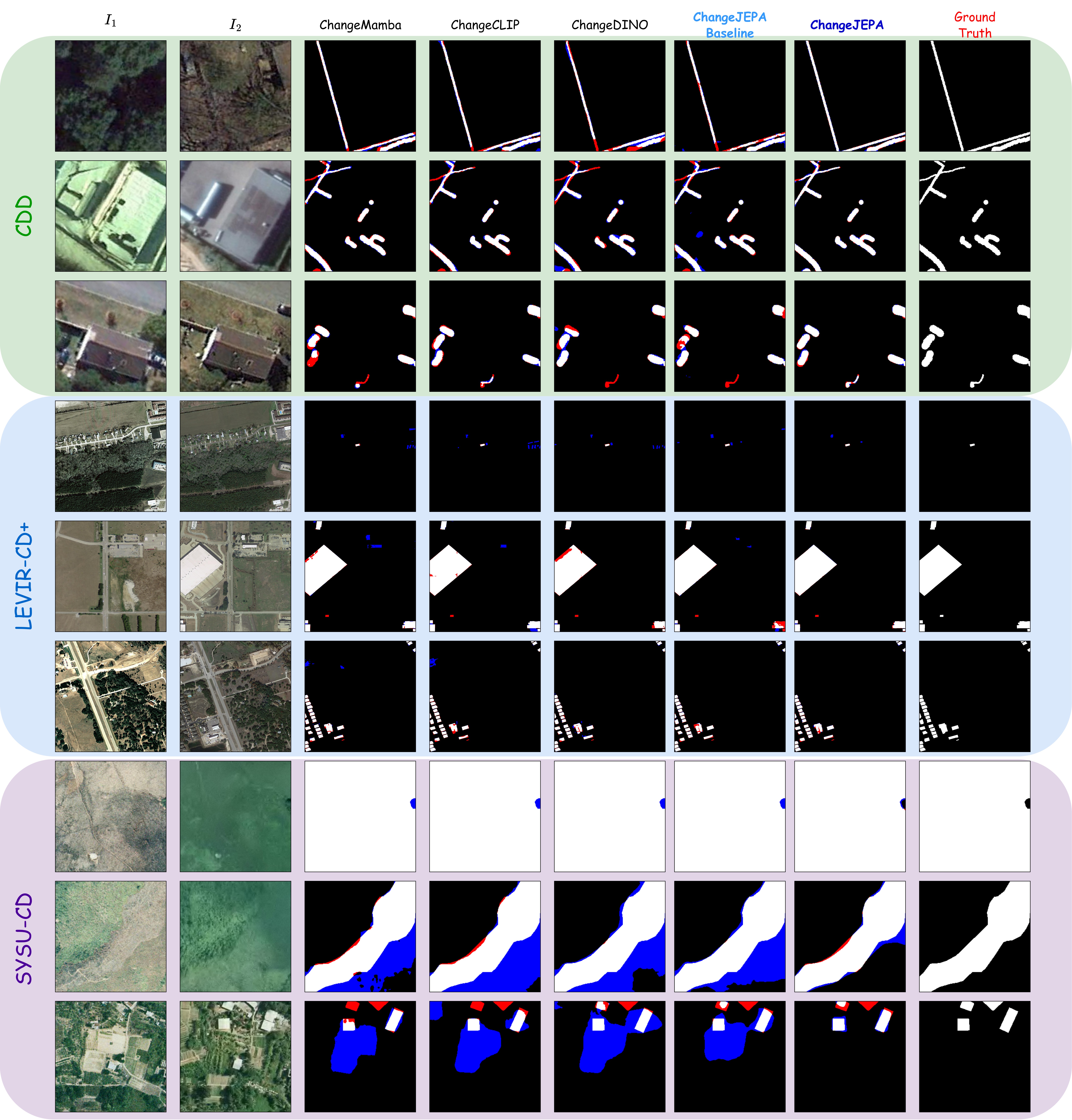}
    \caption{Qualitative comparison on CDD, LEVIR-CD+, and SYSU-CD. From left to
right: pre-change image \(I_1\), post-change image \(I_2\), predictions from
ChangeMamba, ChangeCLIP, ChangeDINO, the decoder-matched ChangeJEPA baseline,
ChangeJEPA, and the ground-truth mask. White, black, blue, and red denote true
positives, true negatives, false positives, and false negatives, respectively.
Across diverse scenes, ChangeJEPA suppresses spurious false positives while
producing cleaner boundaries and more complete predictions for thin structures,
sparse building changes, and large irregular land-cover regions.}
    \label{fig:placeholder}
\end{figure*}

\section{ChangeJEPA Baseline}
\label{sec:changejepa_baseline}

The ChangeJEPA baseline is designed to isolate the contribution of latent
post-state prediction from that of the downstream change-detection architecture.
As illustrated in Fig.~\ref{fig:changejepa_baseline}, it retains the I-JEPA
initialisation, feature dimensions, multi-level skip pathways, evidence
aggregation, MaxViT decoder, and morphological refiner used in ChangeJEPA.
However, it removes the target encoder, post-change masking, bitemporal context
fusion, prior and updated prediction branches, and joint-embedding predictive
objective, yielding a decoder-matched Siamese detector.

The complete pre-change and post-change images are processed independently by
the weight-shared context encoder \(f_{\theta}\), predictor \(g_{\phi}\), and
convolutional projection neck \(\mathcal{N}_{z}\), without masking or
cross-temporal conditioning. The resulting representations \(z_1\) and \(z_2\)
are compared using the same pairwise descriptor
\(\delta(a,b)\) defined in the main paper. This descriptor is
processed by a dedicated MaxViT transformation and CBAM module to produce the
Siamese evidence representation \(E_{\mathrm{S}}\), with the same dimensionality
as the fused evidence in ChangeJEPA.

Intermediate features from transformer blocks \(24\), \(16\), and \(8\) are
similarly compared and passed through the same pixel projectors to construct
the three stage-aligned skip pathways. The resulting evidence is processed by
the unchanged decoder and morphological refiner. The baseline is trained solely
with the change-detection objective \(\mathcal{L}_{\mathrm{cd}}\), without the
predictive warm-up or \(\mathcal{L}_{\mathrm{JEPA}}\).

\section{Datasets}
\label{sec:supp_datasets}

We evaluate ChangeJEPA on three public binary change-detection benchmarks:
LEVIR-CD+, SYSU-CD, and CDD. These datasets cover complementary evaluation
conditions, including building-centred urban change, heterogeneous urban and
land-cover transitions, and pronounced seasonal appearance variation. All
image pairs and ground-truth masks are presented to the network at a spatial
resolution of $256\times256$ pixels. LEVIR-CD+ is partitioned into
non-overlapping $256\times256$ crops, whereas SYSU-CD and CDD are used at
their released $256\times256$ resolution. For every dataset, the experimental splits are
summarised in Table~\ref{tab:dataset_statistics}.

\paragraph{LEVIR-CD+.}
LEVIR-CD+ is an extended version of LEVIR-CD containing 985 pairs of
very-high-resolution RGB images acquired from Google Earth
\cite{chenSpatialTemporalAttentionBasedMethod2020}. Each image has a spatial size of
$1024\times1024$ pixels and a ground sampling distance of $0.5$\,m. The
dataset covers 20 urban regions located in several cities in Texas, USA.
The imagery was acquired between 2002 and 2020. The annotations primarily delineate
building construction and demolition. We use the released split of 637
training and 348 test image pairs. Each image pair and its corresponding mask
are partitioned into a $4\times4$ grid of non-overlapping
$256\times256$ crops, yielding 10,192 training and 5,568 test pairs.

\paragraph{SYSU-CD.}
SYSU-CD contains 20,000 pairs of aerial RGB images acquired over Hong Kong
between 2007 and 2014 \cite{9467555}. Each pair has a spatial size of
$256\times256$ pixels and a ground sampling distance of $0.5$\,m. In contrast
to the building-centred LEVIR-CD+ benchmark, SYSU-CD contains a broader range
of changes, including newly constructed urban buildings, suburban expansion,
construction groundwork, vegetation change, road expansion, and offshore
construction. We split the dataset into 16,000 pairs for training, and 4,000 pairs for testing. No cropping or resizing
is applied.

\paragraph{CDD.}
CDD was introduced to evaluate change detection in season-varying remote
sensing imagery \cite{lebedevCHANGEDETECTIONREMOTE2018}. Its source data
comprise seven season-varying Google Earth image pairs with manually created
annotations and four additional pairs with limited real change into which
target objects were manually introduced. The source imagery has ground
sampling distances ranging from approximately $3$ to $100$\,cm per pixel and
contains objects ranging from vehicles to large constructed structures, as
well as substantial seasonal variation in natural regions. The original paper
reports 16,000 image sets generated as randomly rotated $256\times256$
fragments. We set the training split to 12,998 pairs, and 3,000 pairs for testing. Because the
released samples are already $256\times256$ pixels, no further cropping or
resizing is applied.

\begin{table}[t]
\centering
\caption{Dataset partitions used in all experiments. The training count
denotes the number of samples used for optimisation. All samples
presented to the network are $256\times256$ bitemporal pairs.}
\label{tab:dataset_statistics}
\setlength{\tabcolsep}{5.0pt}
\renewcommand{\arraystretch}{1.10}
\small
\begin{tabular}{@{}lccc@{}}
\toprule
\textbf{Dataset}
& \textbf{Released sample size}
& \textbf{\# Train}
& \textbf{\# Test} \\
\midrule
LEVIR-CD+
& $1024\times1024$
& 10,192
& 5,568 \\
SYSU-CD
& $256\times256$
& 16,000
& 4,000 \\
CDD
& $256\times256$
& 12,998
& 3,000 \\
\bottomrule
\end{tabular}
\end{table}

\section{Evaluation Metrics}
\label{sec:evaluation_metrics}

Following the standard binary change detection protocol, we evaluate the
predicted change maps using Precision, Recall, F1-score, Intersection over Union
(IoU), and Overall Accuracy (OA). The changed class is treated as the positive
class. Let $\mathrm{TP}$ and $\mathrm{TN}$ denote correctly classified changed
and unchanged pixels, respectively, whilst $\mathrm{FP}$ denotes unchanged
pixels incorrectly predicted as changed and $\mathrm{FN}$ denotes changed
pixels incorrectly predicted as unchanged. All valid test pixels are accumulated
into a single confusion matrix before the metrics are computed.

\paragraph{Precision and Recall.}
Precision measures the reliability of the detected change regions, whereas
Recall measures the proportion of reference change pixels recovered by the
model. They are defined as
\begin{equation}
\mathrm{Precision}
=
\frac{\mathrm{TP}}
     {\mathrm{TP}+\mathrm{FP}},
\qquad
\mathrm{Recall}
=
\frac{\mathrm{TP}}
     {\mathrm{TP}+\mathrm{FN}}.
\end{equation}
A high Precision indicates that few unchanged pixels are falsely identified as
changed, whilst a high Recall indicates that few genuine changes are missed.

\paragraph{F1-score.}
The F1-score is the harmonic mean of Precision and Recall:
\begin{equation}
F_{1}
=
\frac{2\,\mathrm{Precision}\,\mathrm{Recall}}
     {\mathrm{Precision}+\mathrm{Recall}}
=
\frac{2\mathrm{TP}}
     {2\mathrm{TP}+\mathrm{FP}+\mathrm{FN}}.
\end{equation}
It provides a balanced assessment of false detections and missed changes and is
therefore informative when the changed and unchanged classes are imbalanced.

\paragraph{Intersection over Union.}
IoU measures the spatial overlap between the predicted and reference change
regions:
\begin{equation}
\mathrm{IoU}
=
\frac{\mathrm{TP}}
     {\mathrm{TP}+\mathrm{FP}+\mathrm{FN}}.
\end{equation}
Unlike OA, IoU excludes correctly classified unchanged pixels and consequently
provides a stricter measure of change-region agreement.

\paragraph{Overall Accuracy.}
OA measures the proportion of all pixels assigned to the correct class:
\begin{equation}
\mathrm{OA}
=
\frac{\mathrm{TP}+\mathrm{TN}}
     {\mathrm{TP}+\mathrm{TN}+\mathrm{FP}+\mathrm{FN}}.
\end{equation}
Since unchanged pixels commonly dominate remote sensing imagery, a high OA may
be obtained even when some change regions are poorly detected. We therefore
interpret OA together with F1 and IoU rather than in isolation.

A negligible constant is added to each denominator in the implementation to
avoid division by zero. All metrics are reported as percentages, with higher
values indicating better performance.

\section{Implementation Details}
\label{sec:implementation_details}

ChangeJEPA is implemented in PyTorch and trained on a single NVIDIA RTX A6000
GPU with 48~GB of memory. All experiments use $256\times256$ bitemporal RGB
images with a batch size of $12$. The context encoder, target encoder, and
predictor are initialised from the I-JEPA ViT-H/16 checkpoint pretrained on
ImageNet-1K for 300 epochs using $448\times448$ images. The pretrained
positional embeddings are bicubically interpolated to the $16\times16$ token
grid, while the remaining modules are initialised from scratch. The target
encoder is updated through an exponential moving average of the context
encoder, with momentum increased linearly from $0.996$ to $1.0$.

Training augmentation includes paired random flipping,
rotation, scaling, and cropping, together with photometric perturbations and
temporal reversal. Identical geometric transformations are applied to both
temporal images and the corresponding mask.

The model is optimised using AdamW with a weight decay of $0.05$ and gradient
clipping at $1.0$. The peak learning rates are $3\times10^{-4}$,
$1\times10^{-4}$, and $1\times10^{-5}$ for the newly initialised modules,
predictor, and context encoder, respectively. The learning rates are linearly
warmed up over the first $15{,}000$ training samples and subsequently decayed
using a cosine schedule. The hidden post-change ratio is sampled from
$\mathcal{U}(0.6,0.8)$. Training begins with $30{,}000$ iterations using only
$\mathcal{L}_{\mathrm{JEPA}}$, followed by joint optimisation for $125{,}000$
iterations on LEVIR-CD+, SYSU-CD and CDD. We use
$\lambda_{\mathrm{prior}}=0.25$, $\lambda_{\mathrm{JEPA}}=1.0$,
$\lambda_{\mathrm{Lov}}=0.5$, and $\alpha=\beta=0.5$ for the Tversky loss.

\noindent\textbf{Inference.}
To ensure deterministic evaluation, we fix the hidden post-change ratio at
$r_{\mathrm{hid}}=0.7$, the midpoint of the training interval, and average the
logits obtained from five block-mask realisations generated using fixed random
seeds. We additionally average predictions from the original, horizontally
flipped, and vertically flipped inputs after mapping them back to their original
orientation. Final labels are obtained by selecting the larger of the unchanged
and changed logits.

\begin{table}[t]
\centering
\caption{Single-view inference without TTA for one
$256\times256$ bi-temporal pair on an NVIDIA RTX A5000.
Latency and FPS are measured in FP32 at batch size 1;
GFLOPs count two operations per multiply--accumulate.}
\label{tab:computational_cost}
\setlength{\tabcolsep}{5pt}
\renewcommand{\arraystretch}{1.08}
\footnotesize
\begin{tabular}{@{}lrrr@{}}
\toprule
Method & GFLOPs $\downarrow$ & Latency (ms) $\downarrow$ & FPS $\uparrow$ \\
\midrule
FC-EF               &     6.19 &   3.06 & 326.4 \\
FC-Siam-Diff        &     8.46 &   4.15 & 241.1 \\
FC-Siam-Conc        &     9.66 &   4.16 & 240.5 \\
IFNet               &   157.86 &  29.30 &  34.1 \\
SNUNet              &     8.46 &   3.78 & 264.7 \\
BIT                 &    17.41 &  13.18 &  75.9 \\
ChangeFormer        &   277.46 &  38.98 &  25.7 \\
ChangeMamba         &   110.05 &  76.24 &  13.1 \\
CDMamba             &   123.08 & 140.68 &   7.1 \\
ChangeRWKV          &    66.42 &  28.95 &  34.5 \\
ChangeCLIP          &    83.37 &  44.35 &  22.5 \\
ChangeDINO          & 1,476.74 & 176.26 &   5.7 \\
\cjb{ChangeJEPA-Baseline} & \cjb{905.07} & \cjb{103.97} & \cjb{9.6} \\
\cj{ChangeJEPA}          & \cj{1,027.10} & \cj{132.80} & \cj{7.5} \\
\bottomrule
\end{tabular}
\end{table}

\begin{table*}[t]
    \centering
    \caption{Endpoint magnitudes and component weights of the controlled
    nuisance perturbation. For component \(c\) at global severity \(s\),
    the effective strength is \(s_c=\min(1,\kappa\,w_c s)\) with dataset intensity
    \(\kappa\), and the operation is interpolated from its identity transformation to
    the reported endpoint. Paired endpoints denote the two possible perturbation directions.}
    \label{tab:illumination_parameters}
    \setlength{\tabcolsep}{4.5pt}
    \renewcommand{\arraystretch}{1.05}
    \small
    \begin{tabular}{@{}llcc@{}}
        \toprule
        \textbf{Family} & \textbf{Component} & \textbf{Endpoint} & \textbf{Weight} \\
        \midrule
        Illumination & Exposure                 & \(\pm1.30\) EV                 & \(0.85\) \\
                     & Gamma                    & \(0.60/1.80\)                  & \(0.70\) \\
                     & Contrast                 & \(0.55/1.60\)                  & \(0.70\) \\
                     & Colour temperature       & \(3500/10000\) K               & \(0.85\) \\
                     & Directional illumination & \(1.00\) EV                    & \(0.85\) \\
                     & Cast shadow              & \(0.65\) attenuation           & \(0.80\) \\
                     & Haze                     & \(0.45\) min.\ transmission    & \(0.55\) \\
        \midrule
        Sensor       & Down/up-sampling         & \(0.45\) min.\ scale           & \(0.55\) \\
                     & Gaussian blur            & \(2.5\) px radius              & \(0.55\) \\
                     & Gaussian noise           & \(\sigma=0.07\)                & \(0.50\) \\
                     & JPEG compression         & quality \(30\)                 & \(0.50\) \\
        \midrule
        Occlusion    & Cloud occlusion          & \(0.60\) opacity               & \(0.55\) \\
        \bottomrule
    \end{tabular}
\end{table*}

\section{Controlled Nuisance Perturbations}
\label{sec:controlled_illumination}

\noindent\textbf{Protocol.}
We examine the sensitivity of ChangeJEPA to controlled appearance
variation on LEVIR-CD+. Four severity levels are
considered, $s\in\{0,0.25,0.5,0.75\}$, denoted by $0\%$, $25\%$,
$50\%$, and $75\%$, respectively. These percentages specify
perturbation strength rather than the proportion of affected
pixels. The $0\%$ condition comprises the original test pairs.
At each non-zero level, only the pre-event image $I_1$ is modified;
the post-event image $I_2$ and reference change mask $Y$ remain
unchanged. The procedure thus alters the visual evidence whilst
preserving the reference labels. No spatial displacement is
introduced, and each severity level constitutes a fixed test set
shared by all methods.

\noindent\textbf{Perturbation Model.}
The compound perturbation comprises successive illumination,
sensor, and occlusion effects. Illumination is varied through
exposure, gamma, contrast, colour temperature, directional
illumination, cast shadow, and haze. Sensor degradation comprises
downsampling and upsampling, Gaussian blur, Gaussian noise, and
JPEG compression, followed by cloud occlusion. The endpoint
magnitudes and component weights are given in
Table~\ref{tab:illumination_parameters}.

For component $c$, the effective strength is
\begin{equation}
    s_c=\min\!\left(1,\,\kappa w_c s\right),
\end{equation}
where $w_c$ is the component weight and $\kappa$ is an intensity
factor, set to $1.3$. Each operation is
interpolated between the identity transformation and its specified
endpoint. Perturbation directions and spatial realisations,
including illumination, shadow, cloud, and noise fields, are
generated from a seed derived from the image filename and global
seed $0$. These are held fixed across severity levels, so that
increasing $s$ strengthens the same perturbation rather than
introducing a new random realisation.

\noindent\textbf{Evaluation.}
All methods are evaluated on identical original and perturbed
test pairs using their best-F1 checkpoints, without retraining
or adaptation. Each checkpoint is held fixed across severity
levels. Inference comprises a single forward pass without
test-time augmentation. ChangeJEPA uses one fixed-seed block
mask with a hidden-patch ratio of $0.7$, retaining the same
masking configuration throughout the experiment.

\noindent\textbf{Results and Interpretation.}
The LEVIR-CD+ results are reported in
Table~\ref{tab:perturbation_levir}. ChangeJEPA achieves the highest
IoU and F1-score among the evaluated methods at each non-zero
severity level. Its IoU is $74.98$, $70.83$, and $53.48$ at
$25\%$, $50\%$, and $75\%$ severity, respectively, compared with
$73.37$, $65.77$, and $47.03$ for the decoder-matched Siamese
baseline. The corresponding advantages are $1.61$, $5.06$, and
$6.45$ percentage points.

Both models deteriorate as severity increases; nevertheless,
the widening margin indicates that ChangeJEPA better preserves
change-region accuracy under the compound degradation considered
here. This finding supports the use of predictive evidence
alongside direct bitemporal comparison when appearance differences
make the interpretation of observed discrepancies more difficult.

Since illumination, sensor degradation, and occlusion are applied
jointly, the experiment assesses their combined effect rather
than the contribution of individual components. The appreciable
loss of accuracy at the highest severity also shows that
ChangeJEPA remains sensitive to substantial degradation of the
pre-event observation. The results therefore support improved
robustness within the tested conditions, without implying
invariance to appearance variation.


\begin{table}[htb]
\centering
\caption{Perturbation robustness on LEVIR-CD+. 25/50/75\% are increasing severities. All methods
use a single forward pass without test-time augmentation. $I$ and $F_1$ denote IoU and F1-score;
all values are reported as percentages. The best and second-best results are highlighted using
\best{boldface} and \second{underlining}, respectively.}
\label{tab:perturbation_levir}
\setlength{\tabcolsep}{3.0pt}
\renewcommand{\arraystretch}{1.08}
\scriptsize
\resizebox{\columnwidth}{!}{%
\begin{tabular}{@{}l|cc|cc|cc@{}}
\toprule
\multirow{2}{*}{\textbf{Method}}
& \multicolumn{2}{c|}{\textbf{25\%}}
& \multicolumn{2}{c|}{\textbf{50\%}}
& \multicolumn{2}{c}{\textbf{75\%}} \\
\cmidrule(lr){2-3}
\cmidrule(lr){4-5}
\cmidrule(l){6-7}
& $I$ & $F_1$ & $I$ & $F_1$ & $I$ & $F_1$ \\
\midrule
ChangeMamba & 65.63 & 79.25 & 53.11 & 69.38 & 40.28 & 57.43 \\
CDMamba     & 63.64 & 77.78 & 50.65 & 67.24 & 43.30 & 60.43 \\
ChangeRWKV  & \second{73.98} & \second{85.05} & \second{65.85} & \second{79.41} & \second{48.83} & \second{65.62} \\
ChangeCLIP  & 70.45 & 82.66 & 54.49 & 70.54 & 41.34 & 58.50 \\
ChangeDINO  & \second{73.98} & 85.04 & 63.51 & 77.69 & 44.23 & 61.33 \\
\cjb{ChangeJEPA-Baseline}
& \cjb{73.37} & \cjb{84.64} & \cjb{65.77} & \cjb{79.35} & \cjb{47.03} & \cjb{63.98} \\
\cj{ChangeJEPA}
& \cj{\best{74.98}} & \cj{\best{85.70}} & \cj{\best{70.83}} & \cj{\best{82.92}} & \cj{\best{53.48}} & \cj{\best{69.69}} \\
\bottomrule
\end{tabular}%
}
\end{table}


\end{document}